\documentclass[format=acmsmall, review=false, screen=true]{acmart}

\usepackage{booktabs} 

\usepackage[ruled]{algorithm2e} 

\SetAlFnt{\small}
\SetAlCapFnt{\small}
\SetAlCapNameFnt{\small}
\SetAlCapHSkip{0pt}
\IncMargin{-\parindent}

\usepackage{epstopdf}
\usepackage{graphicx} 
\graphicspath{ {figs/} }
\usepackage{float}

\usepackage{tabularx}
\usepackage{multirow}
\usepackage{graphicx}
\usepackage{booktabs}
\usepackage{tablefootnote}
\usepackage{amsmath}
\usepackage{amssymb}
\usepackage{comment}
\usepackage{listings}
\usepackage{xcolor}

\usepackage{caption}
\usepackage{subcaption}

\usepackage{dblfnote}
\DFNalwaysdouble 

\acmJournal{CSUR}
\acmVolume{0}
\acmNumber{0}
\acmArticle{0}
\acmYear{2018}

\setcopyright{acmcopyright}

\begin{document}
\title[Toolflows for Mapping CNNs on FPGAs: A Survey and Future Directions]{Toolflows for Mapping Convolutional Neural Networks on FPGAs: A Survey and Future Directions}  

\author{Stylianos I. Venieris, Alexandros Kouris and Christos-Savvas Bouganis}
\orcid{0000-0001-5181-6251}
\affiliation{%
  \institution{Imperial College London}
  \city{London}
  \country{UK}
 }

\begin{abstract}
In the past decade, Convolutional Neural Networks (CNNs) have demonstrated state-of-the-art performance in various Artificial Intelligence tasks. To accelerate the experimentation and development of CNNs, several software frameworks have been released, primarily targeting power-hungry CPUs and GPUs. In this context, reconfigurable hardware in the form of FPGAs constitutes a potential alternative platform that can be integrated in the existing {\color{black}deep learning} ecosystem to provide a tunable balance between performance, power consumption and programmability. In this paper, a survey of the existing CNN-to-FPGA toolflows is presented, comprising a comparative study of their key characteristics which include the supported applications, architectural choices, design space exploration methods and achieved performance. Moreover, major challenges and objectives introduced by the latest trends in CNN algorithmic research are identified and presented. Finally, a uniform evaluation methodology is proposed, aiming at the comprehensive, complete and in-depth evaluation of CNN-to-FPGA toolflows.
\end{abstract}

%
%


%
%


\thanks{The support of the EPSRC Centre for Doctoral Training in High Performance Embedded and Distributed Systems (HiPEDS, Grant Reference EP/L016796/1) is gratefully acknowledged.
Authors' addresses: Department of Electrical and Electronic Engineering, Imperial College London, SW7 2AZ, London, UK; emails: \{stylianos.venieris10, a.kouris16, christos-savvas.bouganis\}@imperial.ac.uk.
}

\maketitle

\renewcommand{\shortauthors}{S. I. Venieris, A. Kouris and C. S. Bouganis}

\section{Introduction}
\label{intro}

Convolutional Neural Networks (CNNs) \cite{Lecun1998} have demonstrated remarkable performance in \mbox{Artificial} Intelligence (AI) tasks. Being able to achieve high accuracy and frequently outperform traditional AI approaches, CNNs have been employed in a vast range of applications over the last decade, from object detection \cite{Ren_2017}\cite{Liu2016ssd} and classification \cite{Simonyan14c}\cite{Szegedy_2016} to drone navigation \cite{drone2017iros} and autonomous driving \cite{Chen2015deepdrive}\cite{Badrinarayanan_2017}. While becoming the state-of-the-art algorithm in AI fields such as machine vision, CNNs are challenged to deal with tasks of continuously increasing complexity. This leads to the design of deeper, more expressive networks at the expense of an increase in computational and memory requirements. 

Several software libraries and frameworks have been developed to facilitate the {\color{black}deep learning} community with the fast development and high-performance execution of CNNs. Toolflows such as Caffe\footnote{http://caffe.berkeleyvision.org/}, Torch\footnote{http://torch.ch/} and Theano\footnote{http://deeplearning.net/software/theano/}, {\color{black}and more recently Caffe2 \footnote{https://caffe2.ai/}, PyTorch\footnote{http://pytorch.org/}, TensorFlow\footnote{https://www.tensorflow.org/}, MXNet\footnote{https://mxnet.apache.org/}, CoreML\footnote{https://developer.apple.com/documentation/coreml}, CNTK\footnote{https://www.microsoft.com/en-us/cognitive-toolkit/} and TensorRT\footnote{https://developer.nvidia.com/tensorrt},} aim to increase the productivity of CNN developers by providing high-level APIs together with high-performance execution of models on power-costly multi-core CPUs, GPUs and DSPs, {\color{black}or on specialised ASICs \cite{Jouppi2017}}. In this context, FPGAs stand as a {\color{black}promising} alternative target platform that can bridge the gap between power-hungry programmable architectures and fixed-function power-efficient ASICs. The reconfiguration capabilities of FPGAs could allow the generation of high-performance, low-power hardware mappings of CNNs that can be configured to meet system-level requirements such as throughput, latency and power in diverse environments, from embedded systems to data centres.

In the last few years, High-Level Synthesis (HLS) tools have demonstrated considerable progress in generating FPGA-based hardware designs from a high level of abstraction \cite{Inggs_2014}. Existing tools such as Xilinx's Vivado HLS, Intel FPGA OpenCL SDK, Maxeler's MaxCompiler {\color{black}and LegUp \cite{Canis_2013}} employ commonly used programming languages such as C, C++, OpenCL and Java in order to facilitate the development of functionally correct hardware designs. Nevertheless, the existing HLS tools aim to yield an efficient design based on the mapping and scheduling of low-level primitive operations, leading to a large design space that does not take into account the inherent structure of the application domain. CNN workloads comprise a well-defined structure consisting of layers, with each layer having a predefined parametrisation. The highly structured nature of CNN workloads enables the development of automated domain-specific frameworks that are tailored to CNNs. Such design tools could represent design points along the most important dimensions of CNNs, by capturing crucial application-level parameters such as the topology of the CNN and the types and configurations of the layers, and map them to architectural parameters.

Currently, various systematic approaches towards the direction of automated mapping of CNNs to FPGAs have been presented. Table \ref{frameworks_table_1} lists the published CNN-to-FPGA toolflows in chronological order. Using the proposed frameworks, an optimised FPGA-based accelerator can be generated, given a CNN-FPGA pair. The integration of this class of accelerator generators in the existing deep learning software frameworks would enable the user community to obtain customised hardware implementations of CNNs, without requiring any hardware design expertise, and thus would enhance the integrability of FPGAs within the deep learning ecosystem.

\begin{table}[t]
\centering
\vspace{-0.2cm}
\caption{CNN-to-FPGA Toolflows 
}
\vspace{-0.25cm}
\label{frameworks_table_1}
\resizebox{0.65\linewidth}{!}{%
	\begin{tabular}{l l r l}
		\toprule
		\multirow{1}{*}{Toolflow Name} & \multicolumn{1}{l}{\multirow{1}{*}{Interface}} & \multicolumn{1}{c}{\multirow{1}{*}{Year}}        \\ 
		\midrule 
		
		fpgaConvNet \cite{Venieris_2016}\cite{Venieris_2017}\cite{Venieris_2017b}\cite{Venieris_2017c} & \begin{tabular}[t]{@{}l@{}} Caffe \& Torch \end{tabular}                                            & May 2016    \\
		
		DeepBurning \cite{Wang_2016}  & Caffe     & June 2016    \\
		
		Angel-Eye \cite{Qiu_2016}\cite{Guo_2016}\cite{Guo_2018}   & \begin{tabular}[t]{@{}l@{}} {\color{black}Caffe} \\ 
		\end{tabular}  & July 2016     \\
		
		ALAMO \cite{Yufei_Ma_2016}\cite{Ma_2017}\cite{Ma_2017b}\cite{Yufei_Ma_2017}\cite{Ma_2018} 			& \begin{tabular}[t]{@{}l@{}} {\color{black}Caffe} 
		\end{tabular} & August 2016 \\
		
		{\color{black} \textsc{Haddoc2} \cite{Abdelouahab_2016}\cite{Abdelouahab_2017}} &{\color{black} Caffe} & {\color{black}September 2016}\\
		
		\textsc{DnnWeaver} \cite{Sharma2016DnnweaverFH}\cite{Sharma_2016}   & Caffe & October 2016 \\ 
		
		Caffeine \cite{Zhang_2016} & Caffe & November 2016 \\
		
		{\color{black}AutoCodeGen \cite{Zhiqiang_Liu_2016}} & {\color{black} Proprietary Input Format} & {\color{black}December 2016} \\
		
		\textsc{Finn} \cite{Umuroglu_2017}\cite{Fraser_2017} & Theano & February 2017 \\
		
		FP-DNN \cite{Guan2017} & TensorFlow & May 2017 \\
		
		{\color{black}Snowflake \cite{Gokhale2017}\cite{Chang_2017}} & {\color{black} Torch} & {\color{black}May 2017} \\
		
		SysArrayAccel \cite{Wei2017} & C Program & June 2017 \\
		
		{\color{black}FFTCodeGen \cite{fft2017fpga}\cite{fftcodegen2017rpt}\cite{fft2017reconfig}\cite{fft2018fpga}} & {\color{black}Proprietary Input Format} & {\color{black}December 2017} \\
		
		\bottomrule
	\end{tabular}%
}
\vspace{-0.1cm}
\end{table}


In this paper, a survey of the various CNN-to-FPGA tooflows is presented. For this work, we consider as a toolflow any developed software that performs direct mapping of any input high-level description of a CNN 
to a hardware architecture that implements the inference computations of the network, under input-specified resource constraints for a target FPGA platform. The paper presents a comparison between these frameworks in terms of supported neural network models, interface, generated hardware architecture, methods used to explore the design space, supported arithmetic precision and performance. Moreover, major challenges introduced by the latest trends in deep learning are identified and possible research directions for automated frameworks are presented. Finally, a benchmark suite together with a uniform evaluation methodology are proposed, aiming at the thorough and in-depth evaluation of CNN-to-FPGA toolflows.

\section{CNN-to-FPGA Toolflow Characteristics}\label{sec2}

In this section, existing toolflows are analysed with respect to their applicability, design methodology and performance. The applicability to an end user is investigated based on the supported neural network models, the input interface and the portability. The design methodology is examined based on the hardware architecture, the design space exploration approach and the arithmetic precision choices. Finally, the performance is analysed based on the reported results of each toolflow.

\subsection{Supported Neural Network Models}\label{sec3}
The application scope of a framework determines the range and type of applications it can target. The majority of the existing toolflows limit their focus on the automated mapping of CNN inference, with
\textsc{Finn} focusing on the more specific field of Binarised Neural Networks (BNNs) \cite{NIPS2016_bnns}. The most common types of layers in a CNN are the convolutional (CONV), nonlinear (NONLIN), pooling (POOL) and fully-connected (FC) layers \cite{Lecun1998}. All existing frameworks support these layers, with ALAMO, DeepBurning, \textsc{DnnWeaver} {\color{black}and AutoCodeGen}  also supporting Local Response Normalisation (NORM) layers \cite{Krizhevsky2012}. Moreover, fpgaConvNet, {\color{black}ALAMO and Snowflake} focus mostly on the feature extractor part of CNNs, including CONV, NONLIN and POOL layers, and offer {\color{black}unoptimised support for FC layers by casting them as CONV layers with $1 \times 1$ kernels. With respect to compound, irregular CNN building blocks, residual blocks \cite{He_2016} are supported by fpgaConvNet, ALAMO and Snowflake, Inception modules \cite{Szegedy2014}\cite{Szegedy_2016} by fpgaConvNet and Snowflake and dense blocks \cite{huang2017densely} by fpgaConvNet.} {\color{black}\textsc{Haddoc2} requires all the weights to be stored on-chip and therefore the supported model size is constrained by the storage resources of the target device.} Currently, DeepBurning and FP-DNN demonstrate the widest range of supported applications by also supporting Recurrent Neural Networks (RNNs) and Long Short-Term Memory (LSTM) networks \cite{lstm1997}.

\subsection{Interface}\label{sec4}
\subsubsection{Input}  \label{sec:inter_IN}
The input interface of an FPGA framework plays a decisive role in its ease-of-use and accessibility to CNN developers. Caffe constitutes the most widely supported front end with support from {\color{black}seven} of the FPGA frameworks, including fpgaConvNet, DeepBurning, Angel-Eye, ALAMO, {\color{black}\textsc{Haddoc2}}, \textsc{DnnWeaver} and Caffeine, due to its structured, protobuf-based\footnote{https://developers.google.com/protocol-buffers/} syntax, the vast availability of pretrained models\footnote{http://caffe.berkeleyvision.org/model\_zoo.html} and the large user community. 
fpgaConvNet {\color{black}and Snowflake} also provide back ends to Torch and FP-DNN has selected TensorFlow as its front end. 
With Theano being the first framework to support BNNs, \textsc{Finn} supports Theano-defined BNNs as its input.

SysArrayAccel, {\color{black}AutoCodeGen and FFTCodeGen} have so far adopted custom front ends. SysArrayAccel uses C programs with embedded pragma directives as its front end and exploits the open-source ROSE\footnote{http://rosecompiler.org/} compiler to capture them. {\color{black}Similarly, AutoCodeGen uses its own proprietary network descriptor, resembling the Caffe syntax. FFTCodeGen employs a custom interface which is based on the YAML\footnote{http://yaml.org/} serialisation framework to specify the CNN model, packaged in a Python 3 wrapper.} The design choice of using custom front ends makes it more difficult to integrate with the existing deep learning toolchains and requires additional infrastructure to make it easily accessible to deep learning practitioners.


\subsubsection{Portability}
A primary characteristic of a CNN-to-FPGA toolflow is the range of supported FPGAs. This feature entails the property of design portability. Portability is defined as the degree to which a toolflow can target FPGA platforms with different specifications. A toolflow with high portability would be able to target (1) devices by multiple vendors and families, (2) different setups such as System-on-Chips (SoCs), host-FPGA servers and standalone FPGA devices as well as \mbox{(3) FPGAs} of different sizes. Moreover, the choice of development tools and level of design, e.g. RTL, vendor-specific HLS or open-source HLS, can affect a toolflow's portability.

Currently, the highest degree of portability has been demonstrated by \textsc{DnnWeaver}. \textsc{DnnWeaver} generates portable RTL in Verilog and has been reported to target both SoCs and server-grade FPGAs from both Xilinx and Intel, including the Xilinx Zynq XC7Z020 SoC and the larger Intel Stratix V GSD5 and Arria 10 GX115. {\color{black}In a similar manner, \textsc{Haddoc2} generates RTL which targets both Intel and Xilinx devices, while AutoCodeGen restricts its scope to RTL targeting Xilinx devices.} {\color{black}fpgaConvNet generates its accelerators in Vivado HLS by Xilinx, while DeepBurning and Angel-Eye use RTL-level design optimised for Xilinx devices.} All three toolflows currently support Xilinx SoCs with results reported on Zynq XC7Z020 and XC7Z045. {\color{black}In a similar manner, Snowflake targets Xilinx SoCs, such as Zynq XC7Z045.} Caffeine is also developed in Vivado HLS and supports server-grade FPGAs with reported results on Kintex UltraScale KU060 and projected results on the larger Virtex 7 VX690T. At the moment, Caffeine's fully automated components target Xilinx devices that support a runnable SDAccel\footnote{https://www.xilinx.com/products/design-tools/software-zone/sdaccel.html} environment and a PCIe interface between the FPGA and a host. {\color{black}FFTCodeGen generates RTL designs in Verilog and targets the Intel Heterogeneous Research Platform (HARP), consisting of tightly coupled CPU and FPGA with shared memory between them. The target FPGA device is Stratix V GXA7, with a 10-core Intel Xeon E5-2600 v2 CPU as a host.}

FP-DNN employs both RTL-level design for its computation engine and Intel OpenCL for interfacing and control logic. In the same direction as Caffeine, FP-DNN targets Intel server-grade FPGAs, with results reported on a Catapult system \cite{Caulfield2016} hosting a Stratix V GSD5 FPGA. Similarly to FP-DNN, SysArrayAccel's hardware is developed in Intel OpenCL with results reported on \mbox{Arria 10 GT115}. \textsc{Finn} generates synthesisable Vivado HLS accelerators and has demonstrated support for the Zynq XC7Z020 and XC7Z045 SoCs as well as the server-grade UltraScale KU115 device in a host-FPGA server setup. Finally, ALAMO's generated RTL designs have demonstrated support for Intel standalone {\color{black}and SoC} platforms by targeting the standalone, high-bandwidth Stratix V GXA7 {\color{black}and the Arria 10 GX115 SoC}.

\begin{figure}
	\centering
	\vspace{-0.25cm}
\begin{minipage}{.58\textwidth}
	\centering
	\includegraphics[width=0.99\textwidth]{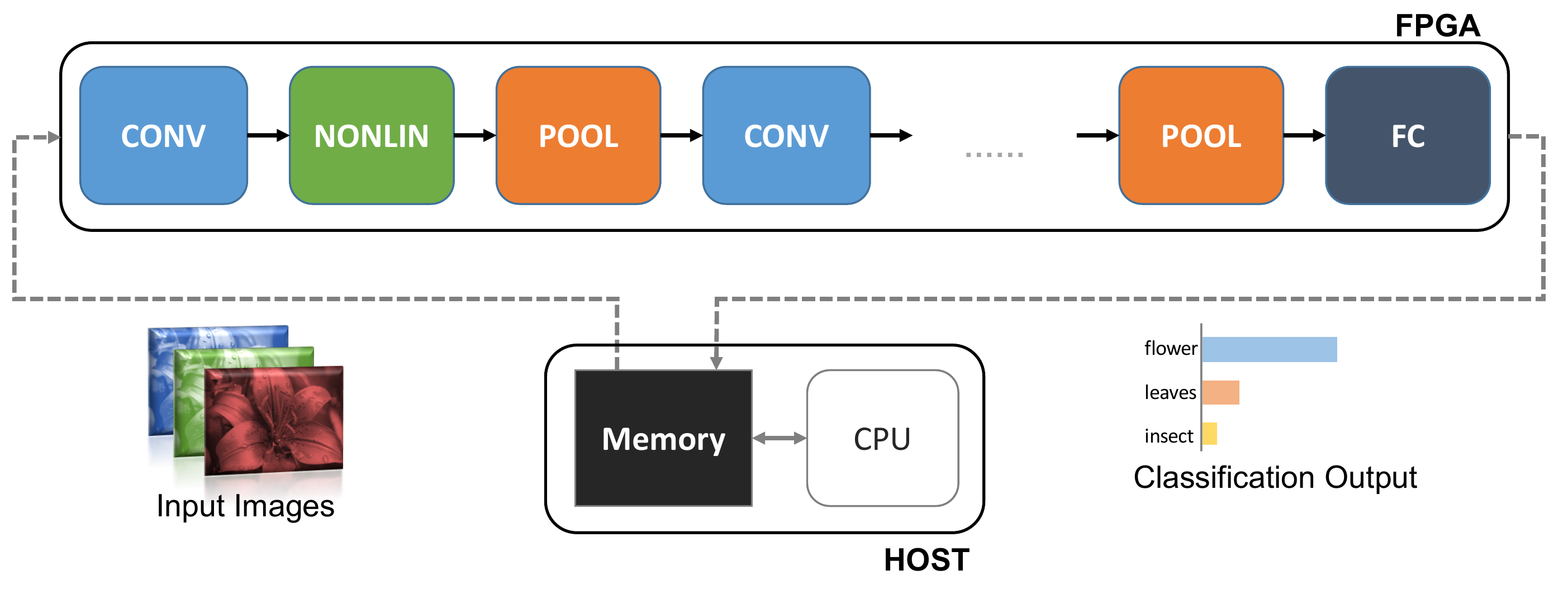}
	\captionof{figure}{Example of a streaming accelerator architecture}
	\label{fig:stream_arch}
\end{minipage}
\begin{minipage}{.39\textwidth}
	\centering
	\includegraphics[width=0.99\textwidth]{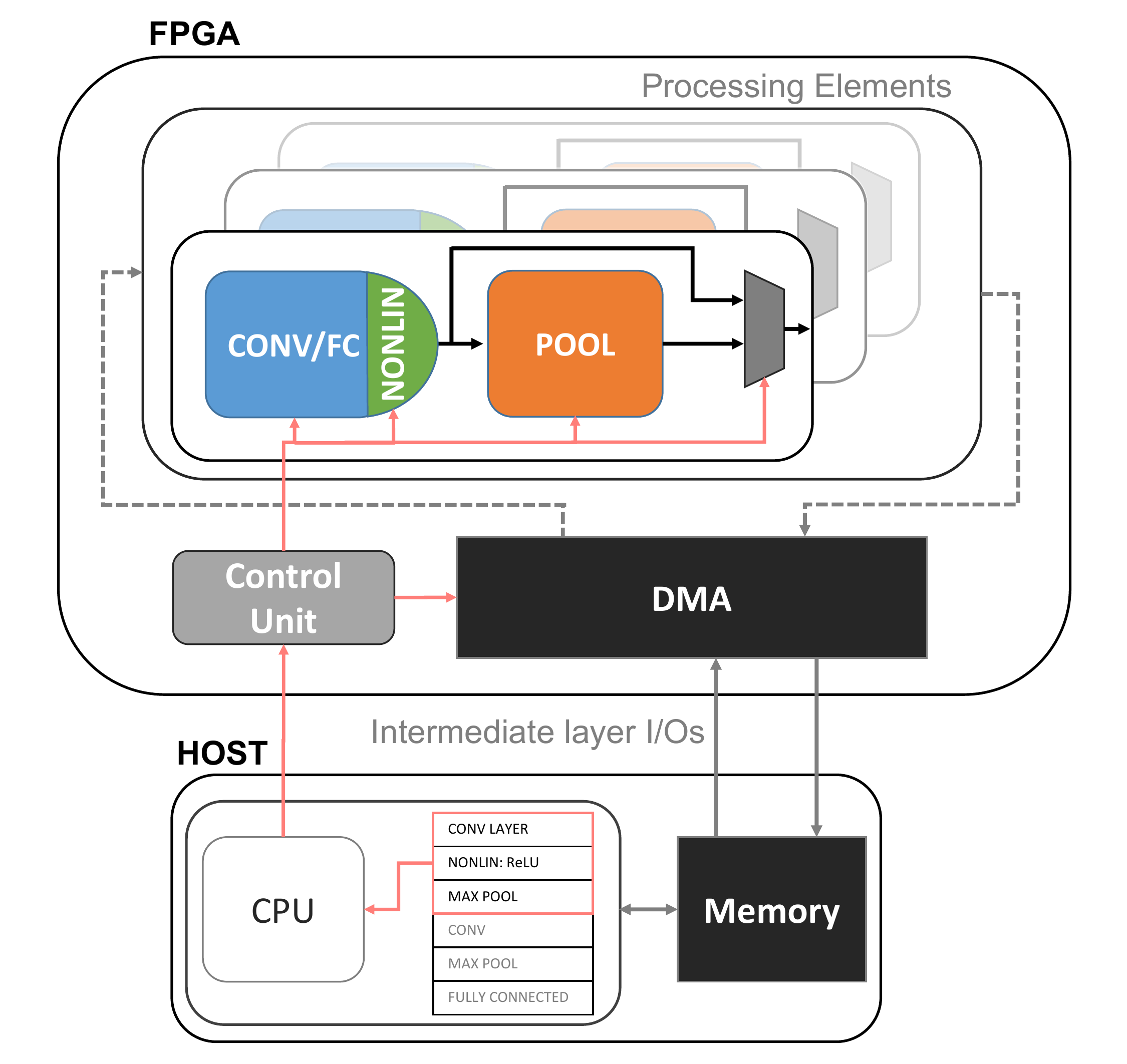}
	\vspace{-0.75cm}
	\captionof{figure}{Example of a single computation \mbox{engine} accelerator}
	\label{fig:comp_eng}
\end{minipage}
\vspace{-0.5cm}
\end{figure}

\subsection{Hardware Architecture}
\label{hw_arch}

The architectures generated by the tools can be taxonomised in two main categories:


\textbf{Streaming architectures:}
A streaming architecture typically consists of one distinct hardware block for each layer of the target CNN, where each block is optimised separately to exploit the parallelism of its layer. All the heterogeneous blocks are chained to form a pipeline as depicted in Fig. \ref{fig:stream_arch}. The data proceed through the different parts of the neural network as they are streamed through the architecture. As a result, this design approach exploits the parallelism between layers by means of pipelining and enables their concurrent execution. Nevertheless, the increased efficiency comes with long compilation times since a new bitstream has to be generated for each CNN.

\textit{1) fpgaConvNet:}
fpgaConvNet employs a streaming architecture which assigns one processing stage per layer. Given a CNN, each layer is mapped to a series of building blocks which are chained together as a coarse pipeline. fpgaConvNet's building blocks include the most commonly utilised components of CNNs, such as convolution and pooling units as well as sliding window structures that provide line-buffering functionality. {\color{black}Moreover, fpgaConvNet employs specialised hardware blocks to map networks with irregular dataflow \cite{Szegedy2014}\cite{Szegedy_2016}\cite{He_2016}\cite{huang2017densely}, including Inception, residual and dense hardware blocks.} The performance-resource trade-off of each instantiated block is tuned separately to meet the needs of each layer in the design space exploration phase. fpgaConvNet supports multi-bitstream designs, where different hardware architectures are responsible for executing different parts of the CNN. Currently, this feature requires the full reconfiguration of the FPGA when data have to enter a new architecture, with the potential for multi-FPGA mappings. 

fpgaConvNet employs a set of strategies to tailor the generated design to the input CNN while respecting the FPGA resources. {\color{black}For latency-sensitive applications, where the time cost of bitstream-level reconfiguration is prohibitive and batch processing cannot be used to amortise it, fpgaConvNet generates a flexible, latency-optimised architecture, which is time-shared to execute different parts of the network by means of soft, run-time reconfiguration of its datapath. Although this latency-driven design approaches the time-shared, single computation engine paradigm, the hardware stages that comprise the architecture are derived and customised based on the structure of the target network and still operate in a streaming manner.} Internally, fpgaConvNet utilises a Synchronous Dataflow (SDF) model \cite{Lee_1987} to represent architectures. With SDF, the processing rates of all blocks in the system are known a priori and therefore a static schedule is generated to drive the datapath.

\textit{2) DeepBurning:}
In a similar approach to fpgaConvNet, DeepBurning's core consists of a library of building blocks that follow the functionality of common neural network components. Currently, the library combines conventional hardware elements such as nonlinear and pooling operators, with  more exotic components such as dropout units \cite{srivastava14a}. Given a network structure, the framework's hardware generator builds the neural network architecture by selecting and instantiating blocks from the library, with the appropriate interconnections between them. To meet the target FPGA resource constraints, each block is parametrised so that it can be time-shared both across layers and across parts of a single layer.

The architecture adopts a run-time, data-driven mechanism where each block executes whenever data are present at its inputs and largely depends on the time-sharing pattern of each block. After the datapath structure and the memory transactions schedule have been determined, the hardware generator creates a centralised control unit, which is responsible for the data movement between the off- and the on-chip memory. Moreover, a dynamic, run-time control approach is adopted by means of dedicated finite state machines that dynamically control the operation of each time-shared block. DeepBurning's dynamic dataflow approach differs from fpgaConvNet's synchronous dataflow scheme in that DeepBurning does not model the data rates of all blocks and thus requires dynamic control logic, rather than generating a static schedule at compile time.

{\color{black}\textit{3) \textsc{Haddoc2}:} \textsc{Haddoc2} generates its architecture by modelling the target CNN as a dataflow graph of actors and directly mapping each actor to a dedicated compute unit. This approach results in the mapping of each layer to a hardware stage, similarly to fpgaConvNet and DeepBurning, with layers executing in parallel in a pipelined manner. The hardware mapping of each layer exploits the full unrolling of its input and output feature maps, and the dot products of convolutions. Unrolling along the three aforementioned dimensions increases the required number of multipliers and on-chip storage, rapidly making the available DSPs and memory of the target FPGA device the limiting factors with respect to the size of CNN that can be mapped. To alleviate the excessive requirement for DSPs, \textsc{Haddoc2} implements all its multipliers solely with logic. Furthermore, since all trained weights are required to be stored on-chip, with off-chip transactions being limited to only the input and output of the network, the weights constitute constant operands for the multipliers. As a result, during synthesis, multiplications with weight values of 0, 1 or powers of 2 are either removed, mapped to direct connections or shift operators respectively. 

With respect to scheduling, \textsc{Haddoc2}'s architecture follows a data-driven approach with the schedule generated statically at compile time. This scheduling method is similar to fpgaConvNet's approach and differs from the dynamic control mechanism of DeepBurning. Nevertheless, in contrast to fpgaConvNet and DeepBurning which support the time-sharing of their resources by means of folding, \textsc{Haddoc2} does not support partial unrolling and, therefore, given a target device, the maximum model size can be quickly bounded either by the available logic or on-chip storage.

}

{\color{black}\textit{4) AutoCodeGen:} 
AutoCodeGen includes parametrised hardware blocks at the layer level, supporting CONV, POOL, NORM and FC layers. CONV blocks consist of convolvers which perform dot-product operations in a fully unrolled manner. The instantiated convolvers are further organised in a tunable number of groups, with input feature maps being shared across all groups. Each convolver group processes the input feature maps with a different set of weights in order to compute independent output feature maps. Within a group, the inputs are parallelised across the convolvers, followed by an adder tree for the reduction of the partial results. FC layers are mapped to compute units, named FCcores, that tunably exploit the input neurons parallelism and can be time-multiplexed. Similarly, POOL blocks exploit the parallelism of output feature maps to a tunable degree. NORM layers are mapped to a fixed hardware block, which employs a piecewise linear approximation scheme for exponential operations and single-precision floating-point arithmetic to minimise precision loss. In contrast to the data-driven control mechanisms of the rest of the toolflows that generate streaming architectures, AutoCodeGen performs the scheduling and control of each hardware block in a distributed manner, with dedicated, local FSMs coordinating the operation of each block.

}

\textit{5) \textsc{Finn}:}
\textsc{Finn} adopts the data-driven paradigm and generates a custom streaming architecture based on a BNN's structure. Given a target BNN, each layer is mapped to a dedicated computation engine and all engines are connected in a pipelined manner. With this design, each computation engine can be configured to meet the requirements of the associated layer and match the processing rate of neighbouring engines. In this manner, the overall architecture is tailored to the particular network. With emphasis placed on BNNs, the computation engines differ from conventional CNN hardware designs and are optimised for the efficient mapping of binarised layers, including dedicated hardware for binarised convolutions, max pooling and batch normalisation \cite{icml2015_ioffe15}. \textsc{Finn} expresses binarised convolutions as matrix-vector operations followed by thresholding. To this end, the integral block of the architecture is the Matrix-Vector-Threshold Unit (MVTU) which is optimised to perform 
the majority of the core binarised operations. In terms of scheduling, \textsc{Finn}'s approach lies closer to fpgaConvNet's synchronous dataflow scheme and farther from DeepBurning's dynamic dataflow, with static schedules generated at compile time. 
Finally, in contrast to fpgaConvNet and DeepBurning {\color{black}and similarly to \textsc{Haddoc2}}, all the binarised weights are required to be stored on-chip, with the external memory transfers focusing only on the input and output of the network, imposing a hard limit to the size of networks that can be addressed.

\textbf{Single computation engines:}
This design approach favours flexibility over customisation. Such an architecture comprises a single computation engine, typically in the form of a systolic array of processing elements or a matrix multiplication unit, that executes the CNN layers sequentially. The control of the hardware and the scheduling of operations is performed by software (Fig. \ref{fig:comp_eng}). 
This design paradigm consists of a fixed architectural template which can be scaled based on the input CNN and the available FPGA resources.
With this scheme, each CNN corresponds to a different sequence of microinstructions that are executable by the hardware. By taking this approach to the extreme, the architecture can be configured and scaled based only on the resources of the target FPGA without targeting a specific CNN and, as a result, after a single compilation, the same bitstream can target many CNNs without the overhead of bitstream-level reconfiguration. Despite the flexibility gains, inefficiencies are introduced due to control mechanisms that resemble those of a processor \cite{Hameed_2010}. Moreover, the one-size-fits-all approach can lead to high variability in the achieved performance across CNNs with different workload characteristics.

\textit{1) Angel-Eye:}
The design principle behind the Angel-Eye framework is based on having a single flexible computation engine which can be programmed and controlled by software. The main computational component is an array of Processing Elements (PEs) with each PE containing a bank of convolvers, an adder tree and an optional pooling path. The input feature maps of a CONV layer are shared across all PEs and each PE processes its inputs with a different set of kernels in order to produce independent output feature maps. Within a PE, the inputs are parallelised across the convolvers, followed by the adder tree that combines partial results to produce the output. {\color{black}Overall, the organisation of Angel-Eye's and AutoCodeGen's hardware for CONV layers are following the same strategy by organising convolvers into groups and tunably unrolling with respect to input and output feature maps.}

The framework's compiler translates the input CNN 
to a sequence of instructions from Angel-Eye's custom instruction set and the computation engine executes the instructions. This process corresponds to the sequential execution of the layers in a time-sharing manner. 
With different CNNs mapped to different instruction sequences, the architecture can be reused to execute various models without recompilation or reconfiguration. In this respect, the hardware design is configured and scaled based only on the available resources of the target device and hence is CNN-independent.

\textit{2) ALAMO:}
{\color{black}In contrast to Angel-Eye, ALAMO customises the generated computation engine to the input CNN. The architecture comprises hardware blocks for POOL, ReLU and NORM layers, together with a 2D array of compute units which is shared between CONV and FC layers. In CONV layers, the array exploits the parallelism within one input feature map and across multiple output feature maps. At each time instant, each row of the array is responsible for one output feature map, with its columns processing different windows of the same input feature map and combining their partial results synergistically. FC layers are mapped on the same hardware block, by casting them as $1 \times 1$ CONV layers. Moreover, ALAMO includes a batch normalisation block and an elementwise adder. These components are employed as complementary to the main blocks, with the elementwise adder used to implement models with irregular dataflow, including residual networks \cite{He_2016}.

Overall, ALAMO's compiler considers the layers that are present in the target CNN and instantiates only the necessary hardware blocks. After the architecture has been generated, the layers are scheduled in a sequential manner. This approach alleviates the problem of allocating resources among different layers of the same type and simplifies the design space to include only the scaling of each hardware block and the scheduling of the layers. The control of the generated accelerator is statically determined at compile time and is encoded as configurations that are loaded sequentially on the accelerator as different parts of the network are executed.}

\textit{3) \textsc{DnnWeaver}:}
\textsc{DnnWeaver}'s hardware is based on a parametrised architectural template. The template comprises an array of coarse Processing Units (PUs). Each PU contains a datapath that includes an array of Processing Elements (PEs) which execute CONV and FC layers, followed by dedicated units for NORM, POOL and NONLIN layers. Within a PU, the CONV and POOL layers are pipelined and their execution is overlapped in order to exploit the parallelism across layers. The computation of output feature maps for CONV and POOL layers and output neurons for FC layers are scheduled across PUs, with PEs exploiting the parallelism between different elements of each output feature map. Generating a specific instance of the template requires trading between the number of PUs and PEs per PU, which resemble the tunable parameters of Angel-Eye's {\color{black}and AutoCodeGen's} architectures. However, in contrast to Angel-Eye which considers only the available resources of the target device, in \textsc{DnnWeaver} this tuning is performed at the design space exploration stage and is tailored to the input CNN and constrained by the resources of the target FPGA, as in the case of ALAMO.

\textit{4) Caffeine:}
Caffeine's hardware consists of a systolic array of PEs that perform multiplication operations. The array offers scalability in implementing convolution operations by exploiting different levels of parallelism, with optional connections between the output of each PE and dedicated blocks for ReLU and POOL layers. Moreover, support for FC layers is achieved by transforming the matrix-vector multiplications of FC layers into batched convolutions and mapping them to the existing convolution structure, which allows the reuse of the exact same hardware for both layers. Given a CNN-FPGA pair, the number of parallel PEs is set after the design space exploration phase, so that the hardware will be tailored to the target CNN.

\textit{5) FP-DNN:}
Drawing from the fact that CONV and FC layers as well as recurrent connections in RNNs and the gate blocks in LSTMs can be converted to matrix multiplications, FP-DNN generates an architecture with a single generic Matrix Multiplication (\textit{MM}) engine as its core. In order to balance the computational resources with the external memory bandwidth, tiling is applied on the input matrices, with the tiles processed in a pipelined manner. The \textit{MM} engine processes the tiles in a vector by vector basis by means of a dot-product unit. The dot-product unit consists of an array of multipliers, which fully unrolls all the multiplications of the dot product, followed by an adder tree. In order to sustain a high utilisation of the computational resources and hide the latency of the off-chip memory, FP-DNN employs double buffering for the transfer of matrix tiles. The \textit{MM} engine is time-shared between layers, with nonlinearities and pooling operations applied by separate hardware prior to writing back intermediate results to the off-chip memory. The on-chip memory is organised as a pool of buffers which can be reused by different data at run time in order to sustain a high utilisation, with the allocation schedule for each buffer handled as part of the design space exploration. Finally, the layer-specific control logic and the interface with the external memory and the host CPU are implemented with OpenCL-based modules.

{\color{black}\textit{6) Snowflake:} 
Snowflake's hardware design employs a hierarchical structure which is designed to be controlled by software. At the top level, the architecture comprises a number of hardware compute clusters, organised as an array of tunable size. Each compute cluster contains four parallel compute units (CUs) with a shared buffer for storing feature maps of the current layer and with each CU consisting of four vector MACC (vMAC) units. Internally, each vMAC includes 16 MACC operators, that process 16-bit operands, together with a private buffer for storing weights of the current layer. In a vMAC, the MACC operators can be configured in two modes, based on the type of parallelism to be exploited. The two modes include either assigning the computation of one output feature map to each MACC operator, exploiting in this way the parallelism with respect to the output feature maps, or assigning the computation of one input feature map to each MAC operator, where the MACC operators collaborate to produce each output feature map by computing partial results. Moreover, each CU also contains a vector max pooling operator (vMAX). Similarly to FP-DNN, double buffering is employed to overlap computation and communication and hide the latency of the external memory transfers.

From an operational perspective, Snowflake is similar to Angel-Eye's programming flow. The target CNN is translated by a custom compiler, named Snowball, into a series of instructions from Snowflake's instruction set and the generated architecture executes the instructions. This process yields the execution of layers in a sequential manner. Moreover, instead of generating a different hardware design for each target CNN, different models are mapped to their own stream of instructions and the architecture can be reused without bitstream-level reconfiguration. In a similar manner to Angel-Eye, the generated hardware is CNN-independent and is scaled based only on the available resources of the target device.

}

\textit{7) SysArrayAccel:}
SysArrayAccel follows Caffeine's approach and adopts a 2D systolic array of PEs to execute all the CONV layers of the target CNN. The main differentiating factor from Caffeine's hardware is that \mbox{SysArrayAccel}'s architecture has been designed so that each PE is only connected locally to its neighbouring PEs. With this approach, SysArrayAccel avoids the need for large multiplexers at the output of each PE, simplifying the routing and achieving high clock frequencies. Each of the two dimensions of the array corresponds to one loop in the CONV layer and each PE performs a configurable number of parallel MACC operations between inputs and weights. The shape of the systolic array can be configured at compile time, so that different degrees of parallelism can be exploited based on the workload characteristics of the target CNN and the available FPGA resources. For the rest of the layers, dedicated hardware blocks are instantiated, with FC layers mapped to a 1D array. Given a CNN-FPGA pair, the selection of loops to be mapped on the systolic array and the shape of the array are selected in the design space exploration phase, in order to optimise the structure of the systolic array for the target CNN.

{\color{black}\textit{8) FFTCodeGen:}
FFTCodeGen differentiates from the rest of the existing toolflows in two main ways. Firstly, FFTCodeGen is optimised to target the heterogeneous Intel HARP platform. In this manner, the framework partitions the CNN workload between the CPU and the FPGA, so that the CONV layers time-share the FPGA device and the rest of the layers are executed in software by the CPU. Secondly, in contrast to the rest of the existing frameworks, FFTCodeGen performs convolutions in the frequency domain by means of an FFT-based algorithm. With this approach, the convolution operations in the space domain are mapped to Hadamard element-by-element products in the frequency domain with decreased computational complexity.

The generated architecture consists of three main components. These comprise 2D FFT and Inverse FFT (IFFT) blocks for transforming feature maps between the space and frequency domains, and a Hadamard-Accumulation (HAC) unit. To perform FFT, FFTCodeGen organises the input feature maps and the kernels as matrices. To support the flexible and tiled FFT-based processing of CONV layers without the need for hardware reconfiguration, FFTCodeGen combines the conventional Overlap-and-Add (OaD) method with the custom Concatenate-and-Pad (CaP) technique. OaD enables the partitioning of the input matrices into tiles of tunable size. The CaP method adds further flexibility by treating the batch size of the network as another dimension of the input feature maps matrix and introduces a tunable folding factor for the batch. The combination of OaD and CaP enable the derivation of a fixed computation engine which can be time-shared among CONV layers with different input feature map sizes, while sustaining high utilisation. In this respect, all tiles of a CONV layer are sequentially fed into the generated accelerator, with double buffering used to hide memory latency, and their partial results are accumulated to produce the output feature maps matrix. Overall, FFTCodeGen uses batch processing to amortise the costs of FFT and IFFT and to enable the CaP method to sustain a high utilisation of the generated accelerator by replacing ineffectual zero-padded operations with useful computations. 

The 2D FFT and IFFT blocks perform $N$-point FFT and IFFT respectively by applying $N$-point 1D FFT on the rows of the input feature maps matrix, followed by an $N$-point 1D FFT on the columns of the transpose of the resulted matrix. The two blocks contain $N$ 1D pipelines each, and share common, tunable folding factors for their rows and columns pipelines. The HAC unit performs elementwise multiplication-accumulation and comprises an array of MACC operators, which is parametrised with respect to its size. FFTCodeGen also comprises software modules for the execution of NONLIN, POOL and FC layers by the CPU. Overall, the processing of CONV layers by the FPGA and the rest of the operations by the CPU are executed in a pipelined manner.

}

\subsection{Design Space Exploration}
\label{param_space_section}
Based on the parametrisation and organisation of its hardware, a toolflow defines a particular architectural design space. Each design point in the design space can be characterised by its performance, including latency and throughput, resource consumption and power efficiency. Typically, a framework would employ a mathematical model of the hardware with the aim to predict how a particular design point performs and investigate how to influence its performance. Design Space Exploration (DSE) refers to the task of traversing the design space and selecting one among the alternative design points based on an application-specific objective. This enables a trade-off between attainable performance and resource distribution and utilisation across the multiple tunable parameters of the architecture, under the resource constraints of the target platform for any given CNN model.

\textbf{Parameter Space.} The proposed architecture of each framework provides different degrees of freedom for customisation, expressed in terms of a set of parameters. fpgaConvNet employs a Synchronous Dataflow (SDF) model \cite{Lee_1987} to capture both the workload and the hardware mapping of CNNs and express them as SDF graphs. Each layer of the input CNN is mapped to a series of coarse hardware blocks, with each block represented as a node of the graph. The architectural space is traversed by applying a set of transformations over the SDF graph representation of the CNN hardware, such as: (1) coarse-grained and (2) fine-grained folding of blocks, (3) graph partitioning with full FPGA reconfiguration and (4) weights reloading. The folding transformations are used to control the degree of time-multiplexing of each block and influence its performance and resource consumption. The FPGA reconfiguration is used to partition the CNN into several subgraphs and effectively change the hardware as the data flow through the CNN, with one optimised hardware design (and bitstream) per subgraph. In this case, batch processing is used to amortise the reconfiguration overhead, with (5) the batch size being a configurable parameter. The weights reloading transformation includes the generation of a single flexible architecture which can be configured at run time to execute different parts of the CNN, by loading different weights from the memory and changing the datapath.

Similarly, \textsc{Finn}'s strategy to maximise performance entails the tailoring of each hardware block along the generated streaming architecture to its layer's workload. To achieve the required performance, the processing rate between the blocks has to be balanced, since the slowest block determines the overall throughput of the system. 
CONV and FC layers are converted to a matrix multiplication between the trained weights and the layer's inputs. With the MVTU being the core computation engine for these operations (Section \ref{hw_arch}), \textsc{Finn} contains a mechanism to fold and time-multiplex the MVTU. Each MVTU in the architecture is compile-time configurable with respect to two parameters: (1) the number of PEs per MVTU and (2) the number of SIMD lanes per PE, which correspond to the \textit{neuron} and \textit{synapse folds} respectively following \textsc{Finn}'s terminology.

DeepBurning's accelerator generation is performed by a hardware generator and a compiler in two steps. As a first step, the hardware generator processes the description of a neural network and creates a baseline architecture. This is achieved by selecting appropriate blocks from DeepBurning's library of neural network components and connecting them as necessary, to create a streaming architecture, as happens in fpgaConvNet and \textsc{Finn}. In the second step, the compiler tunes each block in the architecture so that the accelerator complies with the target FPGA resource constraints. Each block can be configured using (1) temporal folding, where several layers share the same hardware block, and (2) spatial folding, where a single layer is partitioned and all parts are processed by the hardware block in a time-multiplexed manner.

{\color{black}AutoCodeGen instantiates one hardware block per CNN layer. Similarly to \textsc{Finn}, the rate of processing between blocks has to be balanced by tuning the parallelism degree of each hardware block. Each CONV block is compile-time configurable with respect to (1) the number of convolver groups and (2) the number of convolvers per group. Accordingly, each FCcore (Section \ref{hw_arch}) is configurable with respect to (3) the size of the multiplier array and the corresponding adder tree.}

In \textsc{DnnWeaver}, the input CNN is mapped to a dataflow-based intermediate representation, similar to fpgaConvNet. Each node represents an instruction from \textsc{DnnWeaver}'s custom instruction set, with one instruction associated with each layer. The adopted dataflow representation differs from fpgaConvNet's SDF model in that it is utilised to obtain a high-level model of the CNN's workload while fpgaConvNet employs SDF to model both the CNN workload and its hardware mapping. The architectural template is parametrised and tunable with respect to (1) the number of PUs and (2) the number of PEs per PU as described in Section \ref{hw_arch}, as well as with respect to \mbox{(3) the} scheduling of operations. The scheduling is controlled via the tiling factors for each layer's output feature maps, which is processed by each PU, and influences the amount of communication with the off-chip memory.

Caffeine adopts a uniform representation for both the CONV and FC layers which allows the reuse of the same hardware for both layers, as happens in \textsc{Finn}. The design parameters to be optimised include (1) the tiling factors along the three dimensions of the input and output feature maps, (2) the tiling factors of the kernels in CONV layers and (3) the batch size. 

SysArrayAccel interprets CONV layers as nested loops. Analytical performance and resource consumption models have been constructed for the systolic array hardware which are parametrised with respect to (1) the data reuse patterns of the nested loops and (2) the shape of the array. Given a target CNN, different data reuse strategies yield different degrees of parallelism and correspond to selecting two of the nested loops to be mapped on the two dimensions of the systolic array and one loop on the parallel MACC resources of each PE. The shape of the array consists of three parameters which determine the size of each of the two dimensions in the array and the number of parallel MACC units in each PE. SysArrayAccel's tunable parameters enable the exploration along different data reuse strategies and the shaping of the computation engine with three degrees of freedom, in order to traverse the throughput-resource cost space.

{\color{black}ALAMO generates an accelerator by integrating a set of parametrised modules. Depending on the amount of resources of the target device and the distribution of computational workload in the target CNN, different degrees of parallelism (1) within an input feature map and (2) across output feature maps are exploited. FFTCodeGen instantiates $N$-point FFT and IFFT hardware blocks for converting feature maps between the space and frequency domains (1) with $N$ being a design parameter. The two blocks are individually parametrised with respect to (2) the folding factor of each pipeline. The HAC unit (Section \ref{hw_arch}) is also parametrised with respect to (3) the number of MACC operators. Finally, (4) the buffer sizes for feature maps and weights are also tunable.
}

In contrast to the previously described approaches, Angel-Eye's, FP-DNN's {\color{black} and Snowflake's} design principle dictates that the hardware architecture should be independent of the CNN workload. In accordance to this approach, Angel-Eye's generated architecture is parametrised with respect to (1) the number of PEs and (2) the number of convolvers per PE as described in Section \ref{hw_arch} and their values are selected so that the resource utilisation of the target platform is maximised. Similarly, FP-DNN configures its main computation block based only on the available resources of the target platform. As a result, FP-DNN's Matrix Multiplication (\textit{MM}) engine is compile-time configurable with respect to (1) tile size, which is set so that \textit{MM}'s throughput matches the off-chip memory bandwidth of the target platform. Moreover, FP-DNN adopts a resource-sharing strategy for the available on-chip memory resources by organising the on-chip memory as a pool of buffers, with (2) the allocation schedule of each buffer left as a parameter for the DSE. {\color{black}Finally, Snowflake can be scaled at compile time only with respect to (1) the number of compute clusters based on the available resources of the target device, while the number of compute units (CUs) and MACC operators per CU are fixed.}

{\color{black}In a different approach to the rest of the toolflows, \textsc{Haddoc2} captures the input CNN as a dataflow graph of actors and maps each actor to a physical dedicated hardware block, via a process named Direct Hardware Mapping (DHM). With this approach, the architecture is generated deterministically following the exact topology of the network, without configurable parameters.}

\textbf{Design Space Formulation and Search.} The existing FPGA frameworks adopt different levels of analysis for design space exploration which leads to different DSE methods. fpgaConvNet and \textsc{DnnWeaver} cast the DSE as a formal constrained optimisation problem subject to the resource budget of the target FPGA. In each case, the objective function is a mathematical performance model of the hardware, with fpgaConvNet offering either throughput maximisation \cite{Venieris_2016}, latency minimisation \cite{Venieris_2017b} or multiobjective criteria \cite{Venieris_2017c} (such as latency-constrained throughput maximisation) based on the user's needs, while \textsc{DnnWeaver} focuses on throughput maximisation and employs batch processing. Due to the large parameter space that would make a brute-force enumeration intractable, both frameworks employ heuristic search methods in order to obtain a solution to the optimisation problem. \textsc{DnnWeaver} employs a proprietary search algorithm while fpgaConvNet utilises a custom global optimiser based on the Simulated Annealing algorithm \cite{Reeves:1993:MHT:166648}.

Following a different approach, Caffeine bases its DSE on an enhanced version of the roofline model \cite{Williams2009}\cite{Zhang2015}. The refined roofline model yields a better estimate of the effective off-chip memory bandwidth by making it dependent on the burst length of each transfer. In contrast to \textsc{DnnWeaver} and fpgaConvNet, Caffeine's adoption of the higher-level roofline-based modelling leads to a relatively small design space which enables exhaustive enumeration, with the roofline model used to select the design point with the highest throughput subject to the target platform's memory bandwidth and FPGA resources. To limit the latency overhead caused by batch processing, Caffeine converts FC layers to CONV with a method that allows even small batches to reach high throughput.

SysArrayAccel's DSE formulation lies closer to fpgaConvNet's analytical high-level modelling. Emphasis is placed on constructing accurate performance and resource models of the hardware given the selected data reuse patterns and the shape of the systolic array, and casting the DSE as an optimisation problem that aims to maximise throughput. The analytical approach of SysArrayAccel leads to a high-dimensional design space which makes DSE a difficult task. While fpgaConvNet and \textsc{DnnWeaver} employed a global optimiser and a heuristic search algorithm respectively to address this issue, SysArrayAccel applies a number of pruning strategies, including only the consideration of design points that demonstrate high consumption of the FPGA resources, in order to reduce the design space and make an exhaustive enumerative search feasible. As a result, although SysArrayAccel substitutes Caffeine's roofline-based modelling with analytical models, both frameworks employ exhaustive enumeration for the selection of the highest-throughput design point subject to the target off-chip memory bandwidth and FPGA resource constraints.

{\color{black}FFTCodeGen formulates DSE in a manner that combines the analytical approaches of fpgaConvNet, \textsc{DnnWeaver} and SysArrayAccel, with the roofline model of Caffeine. A roofline model is developed as a function of the number of points ($N$) of the FFT, to obtain the value of $N$ that balances the computation-to-communication ratio for the input CNN on the target platform. FFTCodeGen's DSE expresses the computational roof as a function of $N$ and captures the computation-to-communication bound of the target device by means of a single custom metric, named \textit{device coefficient}. This formulation enables the efficient traversal of the high-dimensional design space and differs to the strategies of fpgaConvNet, \textsc{DnnWeaver} and SysArrayAccel to handle large design spaces. After the highest performing $N$ for the target CNN-FPGA pair has been determined, the analytical models are used to obtain the rest of the tunable parameters in a closed form. FFTCodeGen optimises for high throughput, with customisable constraints on the number of points of the FFT and the batch size in order to also support latency-driven applications.}

\textsc{Finn}'s objective is to reach a user-defined throughput. 
The framework's synthesiser module is responsible for determining the values for the folding parameters, using the balancing of the processing rates of all Matrix-Vector-Threshold Units 
as a heuristic. Besides throughput maximisation, \textsc{Finn}'s generated hardware design is also optimised with respect to latency, since no batching of inputs is required. With an approach close to \textsc{Finn}'s, DeepBurning's compiler performs a heuristic search to set the folding parameters of the generated hardware in order to comply with the resource constraints. Similarly to \textsc{Finn}, the generated design runs with a batch size of 1 and hence both throughput and latency are optimised simultaneously. {\color{black}In resemblance to SysArrayAccel and fpgaConvNet, AutoCodeGen employs high-level analytical performance and resource models to set the tunable parameters of each instantiated hardware block, with balancing the processing rates of all hardware blocks as a heuristic, in a similar approach to \textsc{Finn}.}

FP-DNN's mapping strategy focuses on reusing the FPGA resources across layers. With respect to computational resources, the tile size of the single Matrix Multiplication engine is heuristically selected in order to match the off-chip memory bandwidth of the target platform. With respect to memory resources, the allocation schedule of the pool of on-chip buffers is cast as a graph colouring problem which is solved algorithmically, by taking into account the time slots during which the data of each buffer have to remain intact and aiming to find a feasible reuse schedule that maximises buffer utilisation.

{\color{black}ALAMO's DSE focuses on the instantiation of the appropriate hardware blocks, the scaling of each block and the scheduling of layers. The structure of the compute engine is derived based on the topology and layers of the input CNN. After the necassary modules have been instantiated, the compiler's heuristic considers the resource budget of the target FPGA device and determines the unroll factors within an input feature map and across the output feature maps of each layer, in order to scale the 2D array of MACC operators and the POOL block (Section \ref{hw_arch}). ALAMO is designed to operate with a batch size of 1 and therefore throughput and latency are co-optimised.}

In contrast to the rest of the toolflows, Angel-Eye's {\color{black}and Snowflake's} hardware generation are CNN-independent and rely only on the available resources. {\color{black}Each of the two frameworks has a compiler that translates the input CNN to a series of instructions for the accelerator in a heuristic manner.} The DSE process includes the CNN-to-instructions mapping, with throughput maximisation as an objective. When several mappings with equal performance are possible, Angel-Eye's compiler prioritises mappings that minimise the off-chip memory accesses to reduce the bandwidth requirements and power consumption. {\color{black}Moreover, Snowflake's compiler performs optimisations based on the structure of the target CNN, including loop removal, unrolling and rearrangement, and includes a communication load balancing technique to sustain a high utilisation of the compute resources.} Similarly to \textsc{Finn} and DeepBurning, Angel-Eye {\color{black}and Snowflake} are designed to operate with batch size of 1 and hence throughput and latency are co-optimised.

{\color{black}Haddoc2's DHM approach performs a one-to-one mapping between the target network and the generated hardware, without considering the specifications of the target platform. As a result, given an input CNN, the toolflow deterministically generates a hardware design independently of the available resources, and the resulting design is feasible only if it fits within the resource budget of the target device. Moreover, the generated architecture operates with a batch size of 1 and hence is optimised for both throughput and latency.}

\subsection{Arithmetic Precision}

In FPGA-based CNN implementations, data quantisation with few bits has been widely employed. Low-precision fixed-point data representation has been studied to achieve comparable accuracy with high-precision floating-point due to the significant redundancy of the models, while demonstrating a drastic increase in performance \cite{Suda_2016}. The benefits of employing custom-precision arithmetic are manifold, including reducing the external memory bandwidth requirements (and thus decreasing power consumption due to off-chip memory data transfers), minimising the on-chip memory footprint, reducing the resource utilisation by implementing fixed-point arithmetic units and thus leading to better hardware efficiency. 

Based on the observation that significant variation is demonstrated between the dynamic range of data in different layers of the same network, Angel-Eye {\color{black}employs an automated quantisation method to perform dynamic quantisation across layers.} Given a predefined wordlength for the whole network, different scaling, which determines the radix point position, is selected for each layer. Determining the scaling for each layer is formulated as an optimisation problem, solved by a greedy method that minimises the residual error between the network's outputs when fixed-point and floating-point representations are used. {\color{black}After the scaling of each layer has been selected, the quantised, fixed-point weights are fine-tuned by means of a retraining step, to compensate for the accuracy loss due to quantisation.}

ALAMO {\color{black}and AutoCodeGen} allow the wordlength and scaling of each unit to be adjusted at compile time, so that in layers with narrow dynamic range a longer fractional part will be allocated and vice versa. Similarly, \textsc{DnnWeaver} features dedicated bits within each instruction 
(generated by the toolflow's translator module) that dictate whether floating- or fixed-point results should be generated, together with the number of fractional bits and the total bitwidth. Also, in DeepBurning, all components in the hardware library support parametrisable input bitwidth, the value of which is determined by the hardware generator of the toolflow based on the resource constraints.

Caffeine, fpgaConvNet, FP-DNN, SysArrayAccel {\color{black} and FFTCodeGen} provide support for both floating- and fixed-point representations of feature maps and weights. However, a uniform quantisation is applied to all layers in all cases, with fixed wordlength and scaling across them. {\color{black}The same approach is followed by \textsc{Haddoc2}, except that only fixed-point representation is supported.} {\color{black}Snowflake also employs uniform quantisation across all layers, but with a fixed bitwidth of 16 bits.} Finally, \textsc{Finn} consists almost entirely of binary operations as it focuses on BNNs.

\addtolength{\tabcolsep}{0pt} 
	
	\begin{table*}[t]
	    \vspace{-0.2cm}
		\centering
		\caption{Performance Comparison of AlexNet on Zynq and UltraScale Platforms}
        \vspace{-0.25cm}
		\label{alexnet_xilinx_table}
		\resizebox{1\linewidth}{!}{
		\setlength{\tabcolsep}{2pt} 
			\begin{tabular}{l l l l l l l l}
				\toprule
				& \multicolumn{2}{|c|}{fpgaConvNet} & \multicolumn{2}{|c|}{DeepBurning} & \multicolumn{1}{|c|}{\textsc{DnnWeaver}} & \multicolumn{1}{|c|}{Snowflake} &
				\multicolumn{1}{|c}{Caffeine**} \\
				\midrule
				FPGA Platform & Zynq XC7Z020& Zynq XC7Z045 & Zynq XC7Z020 & Zynq XC7Z045 & Zynq XC7Z020 & Zynq XC7Z045 & UltraScale KU060 \\
				Frequency & 125 MHz & 125 MHz & 100 MHz & 100 MHz & 150 MHz & 250 MHz & 200 MHz \\
				Logic Capacity & 53.20 kLUTs & 218.60 kLUTs & 53.20 kLUTs & 218.60 kLUTs & 53.20 kLUTs & 218.60 kLUTs & 331.68 kLUTs \\
				\begin{tabular}[t]{@{}l@{}} 
					DSPs* 
				\end{tabular} & 220 & 900 & 220 & 900 & 220 & 900 & 2760 \\
				On-chip Memory & 0.6 MB & 2.4 MB & 0.6 MB & 2.4 MB & 0.6 MB & 2.4 MB & 4.7 MB \\ Arithmetic Precision & Q8.8 16-bit fixed-point & Q8.8 16-bit fixed-point & 16-bit fixed-point & 16-bit fixed-point & Q3.13 16-bit fixed-point & Q8.8 16-bit fixed-point & Q8.8 16-bit fixed-point \\
				\begin{tabular}[t]{@{}l@{}} 
					Performance \\ 
					(GOp/s) 
				\end{tabular} & \begin{tabular}[t]{@{}l@{}} 
				38.30 (CONV) \\
				-
				\end{tabular} & \begin{tabular}[t]{@{}l@{}} 
				197.40 (CONV) \\
				-
				\end{tabular} & \begin{tabular}[t]{@{}l@{}}
				18.53 (CONV) \\
				15.29 (OVRL)
				\end{tabular}  & \begin{tabular}[t]{@{}l@{}}  
    			108.25 (CONV) \\
    			57.31 (OVRL)
    			\end{tabular} & \begin{tabular}[t]{@{}l@{}} 
    			20.16 (CONV) \\
    			20.51 (OVRL)
    			\end{tabular} & \begin{tabular}[t]{@{}l@{}} 
				120.30 (CONV) \\
				-
				\end{tabular} & \begin{tabular}[t]{@{}l@{}} 
    			163.00 (CONV) \\
    			165.00 (OVRL)
    			\end{tabular} \\
			\begin{tabular}[t]{@{}l@{}} 
				Performance Density \\
				(GOp/s/kLUT)
			\end{tabular} & \begin{tabular}[t]{@{}l@{}}
			$0.72$ (CONV) \\
			- 
			\end{tabular} & \begin{tabular}[t]{@{}l@{}}
			$0.90$ (CONV) \\
			- 
			\end{tabular} & \begin{tabular}[t]{@{}l@{}} 
			$0.35$ (CONV) \\
			$0.29$ (OVRL)
			\end{tabular} & \begin{tabular}[t]{@{}l@{}} 
			$0.49$ (CONV) \\
			$0.26$ (OVRL)
			\end{tabular} & \begin{tabular}[t]{@{}l@{}}
			$0.38$ (CONV) \\
			$0.38$ (OVRL)
			\end{tabular} & \begin{tabular}[t]{@{}l@{}}
			$0.55$ (CONV) \\ 
			- 
			\end{tabular} & \begin{tabular}[t]{@{}l@{}}
			$0.81$ (CONV) \\ 
			$0.83$ (OVRL) 
			\end{tabular} \\ 
			\begin{tabular}[t]{@{}l@{}} 
				Performance Density \\ 
				(GOp/s/DSP) 
			\end{tabular} & \begin{tabular}[t]{@{}l@{}}
			0.1709 (CONV) \\
			- 
			\end{tabular} & \begin{tabular}[t]{@{}l@{}}
			0.2193 (CONV) \\
			- 
			\end{tabular} & \begin{tabular}[t]{@{}l@{}} 
			0.0842 (CONV) \\
			0.0695 (OVRL)
			\end{tabular}& \begin{tabular}[t]{@{}l@{}} 
			0.1203 (CONV) \\
			0.0637 (OVRL)
			\end{tabular} & \begin{tabular}[t]{@{}l@{}}
			0.0916 (CONV) \\
			0.0932 (OVRL)
			\end{tabular} & \begin{tabular}[t]{@{}l@{}}
			0.1336 (CONV) \\
			-
			\end{tabular} & \begin{tabular}[t]{@{}l@{}}
			0.0983 (CONV) \\ 
			0.0996 (OVRL) 
			\end{tabular} \\
			Latency (batch size = 1) & \begin{tabular}[t]{@{}l@{}}
				12.70 ms (CONV) \\
				- 
			\end{tabular} & \begin{tabular}[t]{@{}l@{}}
				8.22 ms (CONV) \\
				- 
			\end{tabular} & \begin{tabular}[t]{@{}l@{}} 
			71.75 ms (CONV) \\
			95.48 ms (OVRL)
			\end{tabular} & \begin{tabular}[t]{@{}l@{}} 
			12.30 ms (CONV) \\
			25.47 ms (OVRL)
			\end{tabular} & \begin{tabular}[t]{@{}l@{}}
			- \\
			-
			\end{tabular} & \begin{tabular}[t]{@{}l@{}}
			9.95 ms (CONV) \\
			-
			\end{tabular} & \begin{tabular}[t]{@{}l@{}}
			- \\
			-
			\end{tabular} \\
			\multicolumn{5}{l}{* 25$\times$18 DSP configurations.} \\
			\multicolumn{8}{l}{** Caffeine's use of SDAccel is reported to have an up-limit of 60\% of the available resources. Therefore, the 60\% is used for the resource-normalised metrics.} \\
			\bottomrule
			\end{tabular}%
			}
\vspace{-0.2cm}
\end{table*}

	
	\begin{table*}[t]
		\centering
		\caption{Performance Comparison of VGG16 on Zynq and UltraScale Platforms}
        \vspace{-0.25cm}
		\label{vgg_xilinx_table}
		\resizebox{\linewidth}{!}{
			\begin{tabular}{l l l l l l}
				\toprule
				& \multicolumn{2}{|c|}{fpgaConvNet} & \multicolumn{1}{|c|}{\textsc{DnnWeaver}} & \multicolumn{1}{|c|}{Angel-Eye} &
				\multicolumn{1}{|c}{Caffeine**}\\
				\midrule
				FPGA Platform & Zynq XC7Z020 & Zynq XC7Z045 & Zynq XC7Z020 & Zynq XC7Z045 & UltraScale KU060 \\
				Frequency & 125 MHz & 125 MHz & 150 MHz & 150 MHz & 200 MHz \\
				Logic Capacity & 53.20 kLUTs & 218.60 kLUTs & 53.20 kLUTs & 218.60 kLUTs & 331.68 kLUTs \\
				\begin{tabular}[t]{@{}l@{}} 
					DSPs* 
				\end{tabular} & 220 & 900 & 220 & 900 & 2760 \\
				On-chip Memory & 0.6 MB & 2.4 MB & 0.6 MB & 2.4 MB & 4.7 MB \\
				Arithmetic Precision & Q8.8 16-bit fixed-point & Q8.8 16-bit fixed-point & Q3.13 16-bit fixed-point & Q8.8 16-bit fixed-point & Q8.8 16-bit fixed-point \\
				\begin{tabular}[t]{@{}l@{}} 
					Performance \\ 
					(GOp/s) 
				\end{tabular} & \begin{tabular}[t]{@{}l@{}} 
				48.53 (CONV) \\
				-
				\end{tabular} & \begin{tabular}[t]{@{}l@{}} 
			    155.81 (CONV) \\
				-
				\end{tabular} & \begin{tabular}[t]{@{}l@{}}
				31.35 (CONV) \\
				31.38 (OVRL)
				\end{tabular}  & \begin{tabular}[t]{@{}l@{}}  
    			187.80 (CONV) \\
    			136.97 (OVRL)
    			\end{tabular} & \begin{tabular}[t]{@{}l@{}}  
    			310.00 (CONV) \\
    			266.00 (OVRL)
    			\end{tabular} \\
			\begin{tabular}[t]{@{}l@{}} 
				Performance Density \\
				(GOp/s/kLUT)
			\end{tabular} & \begin{tabular}[t]{@{}l@{}}
			$0.91$ (CONV) \\
			- 
			\end{tabular} & \begin{tabular}[t]{@{}l@{}}
			$0.71$ (CONV) \\
			- 
			\end{tabular} & \begin{tabular}[t]{@{}l@{}} 
			$0.59$ (CONV) \\
			$0.59$ (OVRL)
			\end{tabular} & \begin{tabular}[t]{@{}l@{}} 
			$0.86$ (CONV) \\
			$0.62$ (OVRL)
			\end{tabular} & \begin{tabular}[t]{@{}l@{}} 
			$1.55$ (CONV) \\ 
			$1.33$ (OVRL) 
			\end{tabular} \\ 
			\begin{tabular}[t]{@{}l@{}} 
				Performance Density \\ 
				(GOp/s/DSP) 
			\end{tabular} & \begin{tabular}[t]{@{}l@{}}
			0.2206 (CONV) \\
			- 
			\end{tabular} & \begin{tabular}[t]{@{}l@{}}
			0.1731 (CONV) \\
			- 
			\end{tabular} & \begin{tabular}[t]{@{}l@{}} 
			0.1425 (CONV) \\
			0.1426 (OVRL)
			\end{tabular} & \begin{tabular}[t]{@{}l@{}} 
			0.2086 (CONV) \\
			0.1522 (OVRL)
			\end{tabular} & \begin{tabular}[t]{@{}l@{}} 
			0.1871 (CONV) \\ 
			0.1606 (OVRL) 
			\end{tabular} \\
			Latency (batch size = 1) & \begin{tabular}[t]{@{}l@{}}
				633.01 ms (CONV) \\
				- 
			\end{tabular} & \begin{tabular}[t]{@{}l@{}}
				249.50 ms (CONV) \\
				- 
			\end{tabular} & \begin{tabular}[t]{@{}l@{}} 
			- \\
			-
			\end{tabular} & \begin{tabular}[t]{@{}l@{}} 
			163.42 ms (CONV) \\
			224.60 ms (OVRL)
			\end{tabular} & \begin{tabular}[t]{@{}l@{}} 
			- \\
			-
			\end{tabular}  \\
			\multicolumn{5}{l}{* 25$\times$18 DSP configurations.} \\
			\multicolumn{6}{l}{** Caffeine's use of SDAccel is reported to have an up-limit of 60\% of the available resources. Therefore, the 60\% is used for the resource-normalised metrics.
			} 
			\\
			\bottomrule
			\end{tabular}%
			}
\vspace{-0.45cm}
\end{table*}

\subsection{Performance}

The most critical characteristic of a CNN-to-FPGA toolflow is the achieved performance of the generated system given a CNN-FPGA pair. An accelerator's primary performance metrics of interest are throughput and latency. 
A tool's 
Quality of Results (QoR) can be evaluated with respect to two factors: (1) comparison with other toolflows for the same CNN-FPGA pair and (2) comparison with hand-tuned accelerators for the same CNN-FPGA pair. Meaningful and fair comparisons across all toolflows would require each toolflow to generate an accelerator for the same CNN targeting the same FPGA device. Nevertheless, the majority of the existing toolflows have not yet been publicly released which does not allow us to obtain results for the same CNN-FPGA benchmarks. At the moment of writing, \textsc{DnnWeaver} has an open-source version\footnote{{\color{black}http://act-lab.org/artifacts/dnnweaver/}} that provides limited support for the Zynq XC7Z020 platform, {\color{black}\textsc{Haddoc2} has been open-sourced\footnote{https://github.com/KamelAbdelouahab/haddoc2}}, \textsc{Finn} has been released in a lightweight version\footnote{{\color{black}https://github.com/Xilinx/BNN-PYNQ}} that targets Xilinx's PYNQ-Z1 board and a set of specific BNNs, fpgaConvNet has a dedicated webpage\footnote{{\color{black}http://cas.ee.ic.ac.uk/people/sv1310/fpgaConvNet.html}} that presents up-to-date benchmarking results on several networks and Angel-Eye is internally used by DeePhi. Due to this fact, the sole feasible method to evaluate each toolflow's achieved performance is by referring to the reported results either in the corresponding publications or by direct communication with the authors. In this study, we combined both approaches in order to collect the presented results.

\addtolength{\tabcolsep}{0pt} 
	
	\begin{table*}[t]
	    \vspace{-0.25cm}
		\centering
		\caption{Performance Comparison of AlexNet on Stratix V and Arria 10}
		\vspace{-0.25cm}
		\label{alexnet_intel_table}
		\resizebox{\linewidth}{!}{
			\begin{tabular}{l l l l l l l}
				\toprule
				& \multicolumn{2}{|c|}{\textsc{DnnWeaver}} & \multicolumn{1}{|c|}{ALAMO} & \multicolumn{1}{|c}{SysArrayAccel} & \multicolumn{1}{|c}{FFTCodeGen} \\
				\midrule
				FPGA Platform & Stratix V SGSD5 & Arria 10 GX115 & Stratix V GXA7 & Arria 10 GT115 & Stratix V GXA7 \\
				Frequency & 200 MHz & 200 MHz & 100 MHz & 239.62 MHz & 200 MHz \\
				Logic Capacity & 172.60 kALMs & 427.20 kALMs & 234.72 kALMs & 427.20 kALMs & 234.72 kALMs \\
				\begin{tabular}[t]{@{}l@{}} 
					DSPs* 
				\end{tabular} & 3180 & 3036 & 512 & 3036 & 512 \\
				On-chip Memory & 4.9 MB & 6.6 MB & 6.25 MB & 6.6 MB & 6.25 MB \\
				Arithmetic Precision & Q3.13 16-bit fixed-point & Q3.13 16-bit fixed-point & Q8.8 16-bit fixed-point & 32-bit floating-point & 16-bit fixed-point \\
				\begin{tabular}[t]{@{}l@{}} 
					Performance \\ 
					(GOp/s) 
				\end{tabular} & \begin{tabular}[t]{@{}l@{}} 
				97.10 (CONV) \\
				97.56 (OVRL)
				\end{tabular} & \begin{tabular}[t]{@{}l@{}} 
				265.36 (CONV) \\
				184.33 (OVRL)
				\end{tabular} & \begin{tabular}[t]{@{}l@{}}
				134.10 (CONV) \\
				114.50 (OVRL)
				\end{tabular}  & \begin{tabular}[t]{@{}l@{}}  
    			406.1 (CONV) \\
    			360.4 (OVRL)
    			\end{tabular} & \begin{tabular}[t]{@{}l@{}}  
    			- \\
    			780.60 (OVRL) 
    			\end{tabular} \\
			\begin{tabular}[t]{@{}l@{}} 
				Performance Density \\
				(GOp/s/kALM)
			\end{tabular} & \begin{tabular}[t]{@{}l@{}}
			$0.56$ (CONV) \\
			$0.56$ (OVRL)
			\end{tabular} & \begin{tabular}[t]{@{}l@{}}
			$0.62$ (CONV) \\
			$0.43$ (OVRL)
			\end{tabular} & \begin{tabular}[t]{@{}l@{}} 
			$0.57$ (CONV) \\
			$0.49$ (OVRL)
			\end{tabular} & \begin{tabular}[t]{@{}l@{}} 
			$0.95$ (CONV) \\
			$0.84$ (OVRL)
			\end{tabular} & \begin{tabular}[t]{@{}l@{}} 
			- \\
			$3.32$ (OVRL)
			\end{tabular} \\ 
			\begin{tabular}[t]{@{}l@{}} 
				Performance Density \\ 
				(GOp/s/DSP) 
			\end{tabular} & \begin{tabular}[t]{@{}l@{}}
			0.0305 (CONV) \\
			0.0307 (OVRL)
			\end{tabular} & \begin{tabular}[t]{@{}l@{}}
			0.0874 (CONV) \\
			0.0607 (OVRL)
			\end{tabular} & \begin{tabular}[t]{@{}l@{}} 
			0.2619 (CONV) \\
			0.2236 (OVRL)
			\end{tabular} & \begin{tabular}[t]{@{}l@{}} 
			0.1337 (CONV) \\
			0.1187 (OVRL)
			\end{tabular} & \begin{tabular}[t]{@{}l@{}} 
			- \\
		    1.5246 (OVRL) 
			\end{tabular} \\
			Latency (batch size = 1) & \begin{tabular}[t]{@{}l@{}}
				- \\
				- 
			\end{tabular} & \begin{tabular}[t]{@{}l@{}}
				- \\
				- 
			\end{tabular} & \begin{tabular}[t]{@{}l@{}} 
			9.92 ms (CONV) \\
			12.75 ms (OVRL)
			\end{tabular} & \begin{tabular}[t]{@{}l@{}} 
			- \\
			4.05 ms (OVRL)
			\end{tabular} & \begin{tabular}[t]{@{}l@{}} 
			- \\
		    -
			\end{tabular} \\
			\multicolumn{5}{l}{* 18$\times$18 DSP configurations.} \\
			\bottomrule
			\end{tabular}%
			}
\vspace{-0.3cm}
\end{table*}

	
	\begin{table*}[t]
		\centering
		\caption{Performance Comparison of VGG16 on Stratix V and Arria 10}
        \vspace{-0.25cm}
		\label{vgg_intel_table}
		\resizebox{\linewidth}{!}{
		\setlength{\tabcolsep}{6pt} 
			\begin{tabular}{l l l l l l l l}
				\toprule
				& \multicolumn{2}{|c|}{\textsc{DnnWeaver}} & \multicolumn{2}{|c|}{\textsc{ALAMO}} & \multicolumn{1}{|c|}{FP-DNN**} & \multicolumn{1}{|c}{SysArrayAccel} & \multicolumn{1}{|c}{FFTCodeGen} \\
				\midrule
				FPGA Platform & Stratix V SGSD5 & Arria 10 GX115 & Stratix V GXA7 & Arria 10 GX115 & Stratix V SGSMD5 & Arria 10 GT115 & Stratix V GXA7 \\
				Frequency & 200 MHz & 200 MHz & 150 MHz & 200 MHz & 150 MHz & 231.85 MHz & 200 MHz \\
				Logic Capacity & 172.60 kALMs & 427.20 kALMs & 234.72 kALMs & 427.20 kALMs & 172.60 kALMs & 427.20 kALMs & 234.72 kALMs \\
				\begin{tabular}[t]{@{}l@{}} 
					DSPs* 
				\end{tabular} & 3180 & 3036 & 512 & 3036 & 3180 & 3036 & 512 \\
				On-chip Memory & 4.9 MB & 6.6 MB & 6.25 MB & 6.6 MB & 4.9 MB & 6.6 MB & 6.25 MB \\
				Arithmetic Precision & Q3.13 16-bit fixed-point & Q3.13 16-bit fixed-point & Q8.8 16-bit fixed-point & Q8.8 16-bit fixed-point & 16-bit fixed-point & 16-bit fixed-point & 16-bit fixed-point \\
				\begin{tabular}[t]{@{}l@{}} 
					Performance \\ 
					(GOp/s) 
				\end{tabular} & \begin{tabular}[t]{@{}l@{}} 
				157.39 (CONV) \\
				157.51 (OVRL)
				\end{tabular} & \begin{tabular}[t]{@{}l@{}} 
				390.02 (CONV) \\
				361.55 (OVRL)
				\end{tabular} & \begin{tabular}[t]{@{}l@{}}
				- \\
				352.24 (OVRL)
				\end{tabular} & \begin{tabular}[t]{@{}l@{}}
				- \\
				720.15 (OVRL)
				\end{tabular} & \begin{tabular}[t]{@{}l@{}}
				- \\
				364.36 (OVRL)
				\end{tabular} & \begin{tabular}[t]{@{}l@{}}
				- \\
				1171.30 (OVRL)
				\end{tabular} & \begin{tabular}[t]{@{}l@{}}
				- \\
				669.10 (OVRL)
				\end{tabular} \\
			\begin{tabular}[t]{@{}l@{}} 
				Performance Density \\
				(GOp/s/kALM)
			\end{tabular} & \begin{tabular}[t]{@{}l@{}}
			$0.91$ (CONV) \\
			$0.91$ (OVRL)
			\end{tabular} & \begin{tabular}[t]{@{}l@{}}
			$0.91$ (CONV) \\
			$0.84$ (OVRL)
			\end{tabular} & \begin{tabular}[t]{@{}l@{}} 
			- \\
			$1.50$ (OVRL)
			\end{tabular} & \begin{tabular}[t]{@{}l@{}} 
			- \\
			$2.74$ (OVRL)
			\end{tabular} & \begin{tabular}[t]{@{}l@{}} 
			- \\
			$2.11$ (OVRL)
			\end{tabular} & \begin{tabular}[t]{@{}l@{}} 
			- \\
			$2.74$ (OVRL)
			\end{tabular} & \begin{tabular}[t]{@{}l@{}} 
			- \\
			$2.85$ (OVRL) 
			\end{tabular} \\ 
			\begin{tabular}[t]{@{}l@{}} 
				Performance Density \\ 
				(GOp/s/DSP) 
			\end{tabular} & \begin{tabular}[t]{@{}l@{}}
			0.0495 (CONV) \\
			0.0495 (OVRL)
			\end{tabular} & \begin{tabular}[t]{@{}l@{}}
			0.1284 (CONV) \\
			0.1191 (OVRL)
			\end{tabular} & \begin{tabular}[t]{@{}l@{}} 
			- \\
			0.6879 (OVRL)
			\end{tabular} & \begin{tabular}[t]{@{}l@{}} 
			- \\
			0.2372 (OVRL)
			\end{tabular} & \begin{tabular}[t]{@{}l@{}} 
			- \\
			0.1145 (OVRL)
			\end{tabular} & \begin{tabular}[t]{@{}l@{}} 
			- \\
			0.3858 (OVRL)
			\end{tabular} & \begin{tabular}[t]{@{}l@{}} 
			- \\
			1.3068 (OVRL)
			\end{tabular} \\
			Latency (batch size = 1) & \begin{tabular}[t]{@{}l@{}}
				- \\
				- 
			\end{tabular} & \begin{tabular}[t]{@{}l@{}}
				- \\
				- 
			\end{tabular} & \begin{tabular}[t]{@{}l@{}} 
			- \\
			87.87 ms
			\end{tabular} & \begin{tabular}[t]{@{}l@{}} 
			- \\
			42.98 ms (OVRL)
			\end{tabular} & \begin{tabular}[t]{@{}l@{}} 
			- \\
			-
			\end{tabular} & \begin{tabular}[t]{@{}l@{}} 
			- \\
			26.85 ms (OVRL)
			\end{tabular} & \begin{tabular}[t]{@{}l@{}} 
			- \\
			-
			\end{tabular} \\
			\multicolumn{5}{l}{* 18$\times$18 DSP configurations.} \\
			\multicolumn{5}{l}{** FP-DNN maps VGG19 on Stratix V.} \\
			\bottomrule
			\end{tabular}%
		}
\vspace{-0.5cm}
\end{table*}

In this section, a performance comparison is presented with the aim to depict an as much as possible well-rounded view of the strengths and weaknesses of each toolflow and draw conclusions about the different mapping strategies. Our evaluation methodology consists of two components: (1) in order to conduct a fair and meaningful evaluation, we perform direct comparisons only between tools that have mapped the same CNN model on the same FPGA device and (2) we assess the quality of the automatically generated designs by comparing with the current state-of-the-art, hand-tuned designs for the same CNN-FPGA pairs.

So far, results have been reported on a variety of CNN models, with different tools selecting different benchmarks and devices. Our evaluation is focused on the most commonly mapped AlexNet and VGG16 networks, {\color{black}with a number of additional comparisons on LeNet-5, CIFAR-10, GoogLeNet and ResNet-152.} Detailed results for both the feature extractors (CONV) and the feature extractors followed by classifiers (OVRL)\footnote{The complete performance results were obtained by contacting the authors.} are listed in Tables \ref{alexnet_xilinx_table} and \ref{vgg_xilinx_table} for AlexNet and VGG16 respectively on Zynq and UltraScale platforms and in Tables \ref{alexnet_intel_table} and \ref{vgg_intel_table} for AlexNet and VGG16 respectively on Stratix V and Arria 10 platforms. Resource-normalised metrics\footnote{Results are normalised over the available resources of the target device.} are also included since, despite their limitations which are discussed in Section \ref{eval_metrics_sec}, they constitute the current literature standard metric for CNN accelerator comparisons across different devices. For platforms from the same FPGA family and vendor, normalisation with respect to LUTs and ALMs can be used. For heterogeneous platforms, normalisation with DSPs is employed.


\begin{figure*}[t]
    \centering
    \includegraphics[trim={0cm 4cm 0cm 4cm},clip,width=0.7\textwidth]{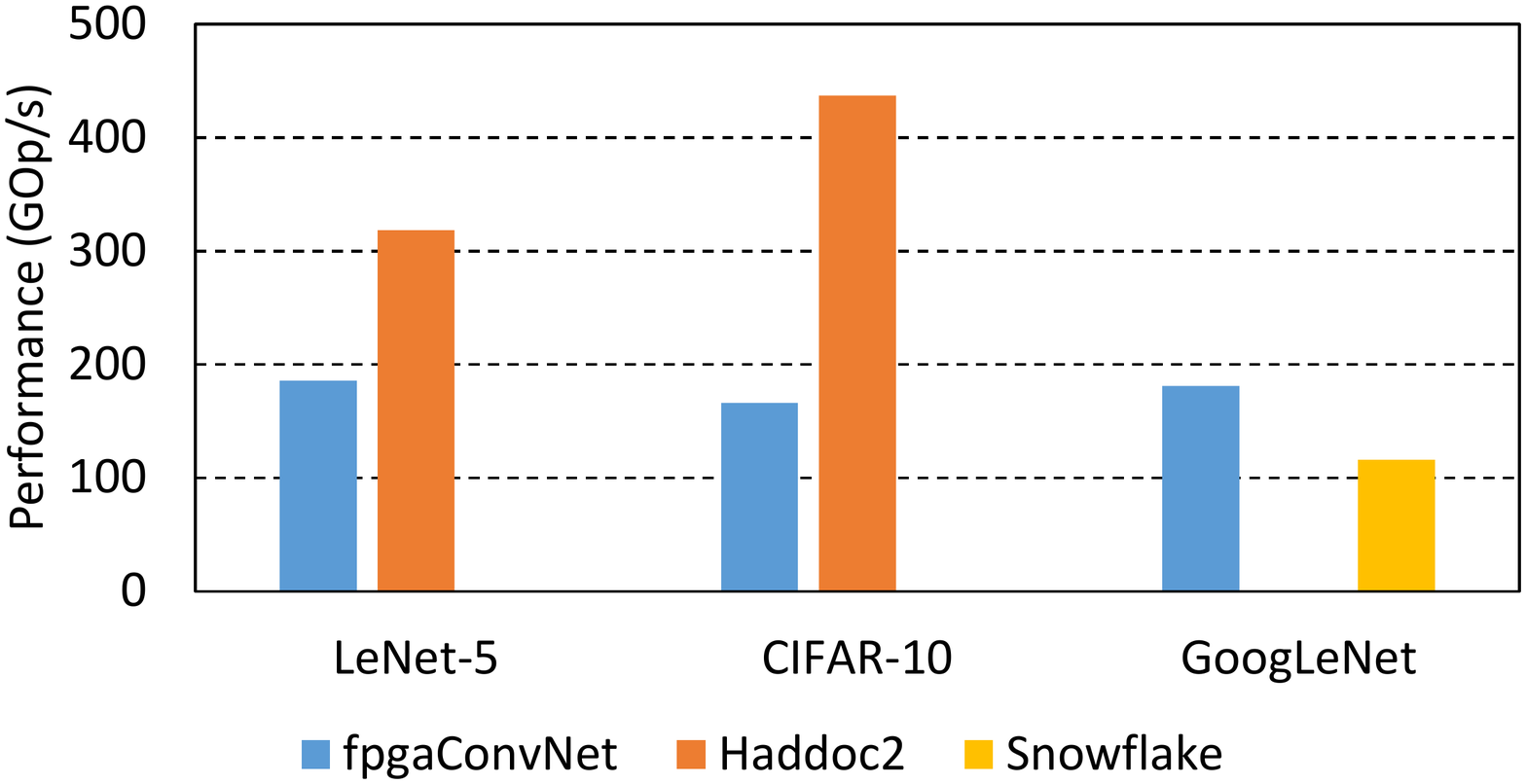}
    \vspace{-0.3cm}
    \caption{Comparison on mapping LeNet-5, CIFAR-10 and GoogLeNet on Zynq XC7Z045}
    \label{fig:zynq_others}
    \vspace{-0.5cm}
\end{figure*}



\begin{figure}
	\centering
	\begin{subfigure}{.48\linewidth}
	    \centering
	    {\includegraphics[trim={1.25cm 2cm 1cm 2cm},clip,width=1\textwidth]{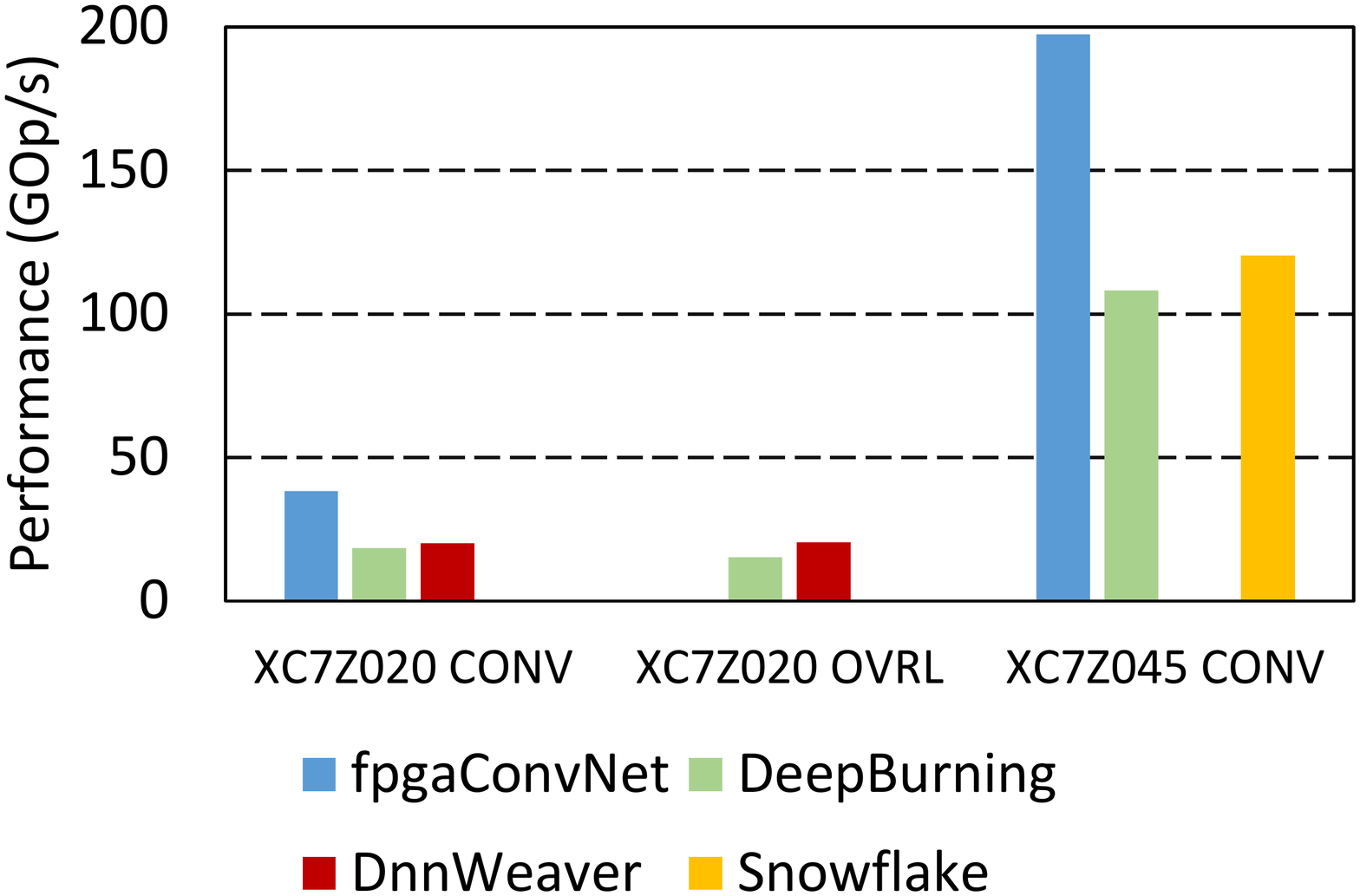}}
			\vspace{-0.6cm}
			\caption{AlexNet}
			\label{fig:alexnet_tools}
	\end{subfigure}
	\begin{subfigure}{.48\linewidth}
		\centering
		{\includegraphics[trim={1cm 1.8cm 1cm 3cm},clip,width=1\textwidth]{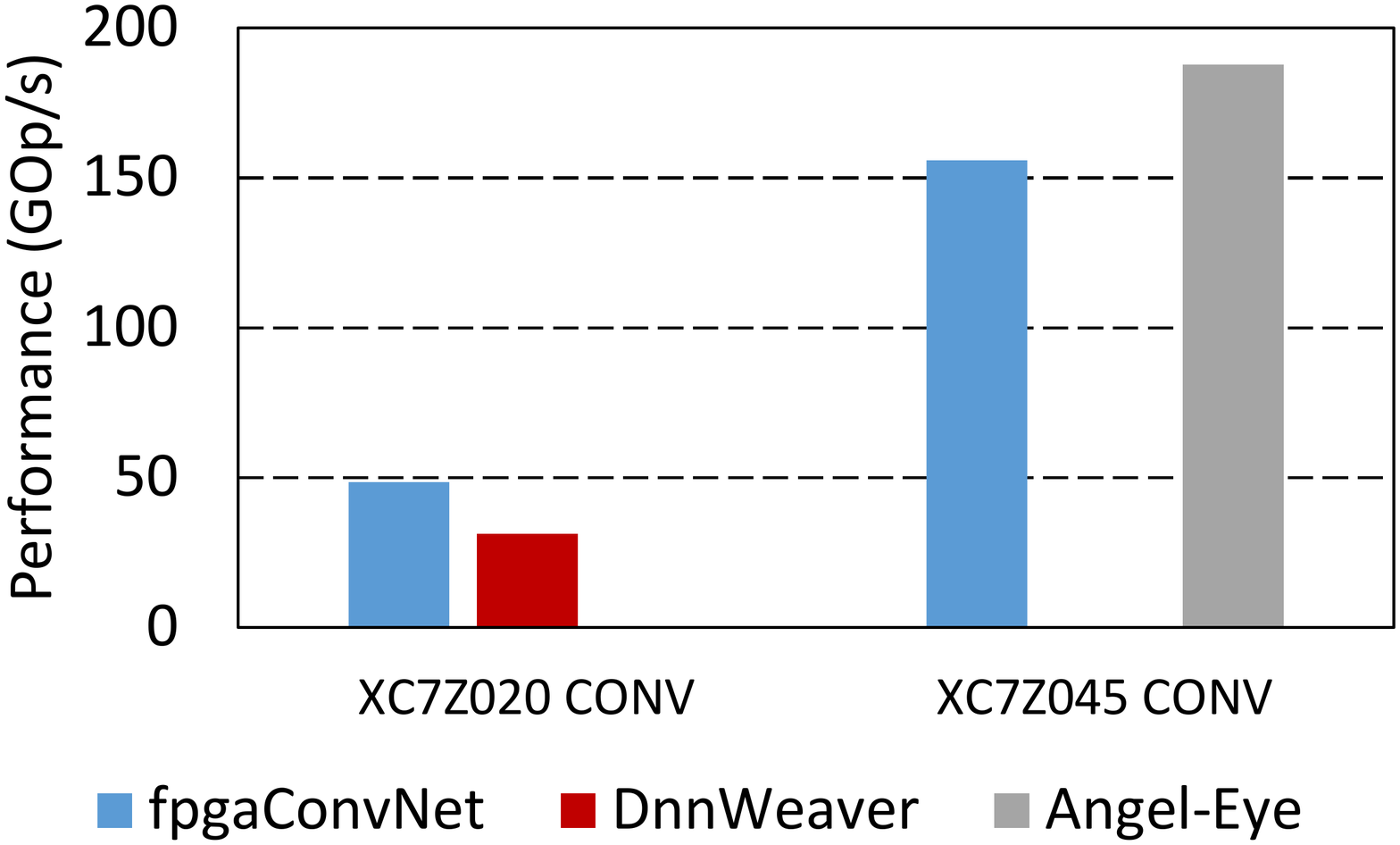}}
		\vspace{-0.5cm}
		\caption{VGG16}
		\label{fig:vgg16_tools}
	\end{subfigure}
	\vspace{-0.4cm}
	\caption{Comparison targeting Zynq platforms} 
	\label{fig:zynq_comp}
	\vspace{-0.55cm}
\end{figure}

\textbf{Comparison between Toolflows.}
Fig. \ref{fig:alexnet_tools} and \ref{fig:vgg16_tools} present comparisons of toolflows for the mapping of AlexNet and VGG16 on Zynq platforms. fpgaConvNet, DeepBurning and \textsc{DnnWeaver} mapped AlexNet on the resource-limited Zynq XC7Z020 platform (Fig. \ref{fig:alexnet_tools}). With respect to the feature extractor, fpgaConvNet achieves a throughput of $38.30$ GOp/s and outperforms DeepBurning and \textsc{DnnWeaver} by $2.06 \times$ and $1.9 \times$ respectively, while for the whole AlexNet, \textsc{DnnWeaver} reaches $1.34 \times$ higher throughput than DeepBurning. With respect to latency, fpgaConvNet's latency-driven methodology yields a $1.37 \times$ lower latency than DeepBurning for AlexNet's feature extractor. \textsc{DnnWeaver} has been optimised for high-throughput applications and requires batch processing to achieve high performance. Therefore, \textsc{DnnWeaver}'s latency has not been considered. When targeting the resource-richer Zynq XC7Z045, fpgaConvNet achieves $1.82 \times$ higher throughput and $1.49 \times$ lower latency compared to DeepBurning, demonstrating a similar trend to AlexNet on Zynq XC7Z020. {\color{black}Compared to Snowflake, fpgaConvNet reaches $1.64 \times$ higher throughput and $1.21\times$ lower latency, with Snowflake achieving $1.11\times$ higher throughput than DeepBurning.} With respect to mapping VGG16 on Zynq XC7Z020 (Fig. \ref{fig:vgg16_tools}), fpgaConvNet achieves $1.22 \times$ higher throughput than \textsc{DnnWeaver}. The gap between the two toolflows is small and possibly due to the finer exploration method of fpgaConvNet.



Both fpgaConvNet and Angel-Eye have mapped VGG16 on Zynq XC7Z045 (Fig. \ref{fig:vgg16_tools}). Angel-Eye has achieved the current state-of-the-art performance of VGG16 on Zynq XC7Z045 with $1.20 \times$ higher throughput than fpgaConvNet for the feature extractor and $136.97$ GOp/s for the whole network. Moreover, Angel-Eye achieves $1.52 \times$ lower latency than fpgaConvNet. 

{\color{black}fpgaConvNet and \textsc{Haddoc2} have both generated accelerators for the low-end LeNet-5 and CIFAR-10 on Zynq XC7Z045 (Fig. \ref{fig:zynq_others}). In these two cases, \textsc{Haddoc2} achieves $1.71\times$ on LeNet-5 and $2.63\times$ on CIFAR-10 higher throughput than fpgaConvNet, while using 3-bit and 6-bit bitwidth respectively for the two networks compared to the 16-bit representation of fpgaConvNet.}

\begin{figure*}[t]
    \centering
    \includegraphics[trim={2cm 4cm 0cm 4cm},clip,width=0.8\textwidth]{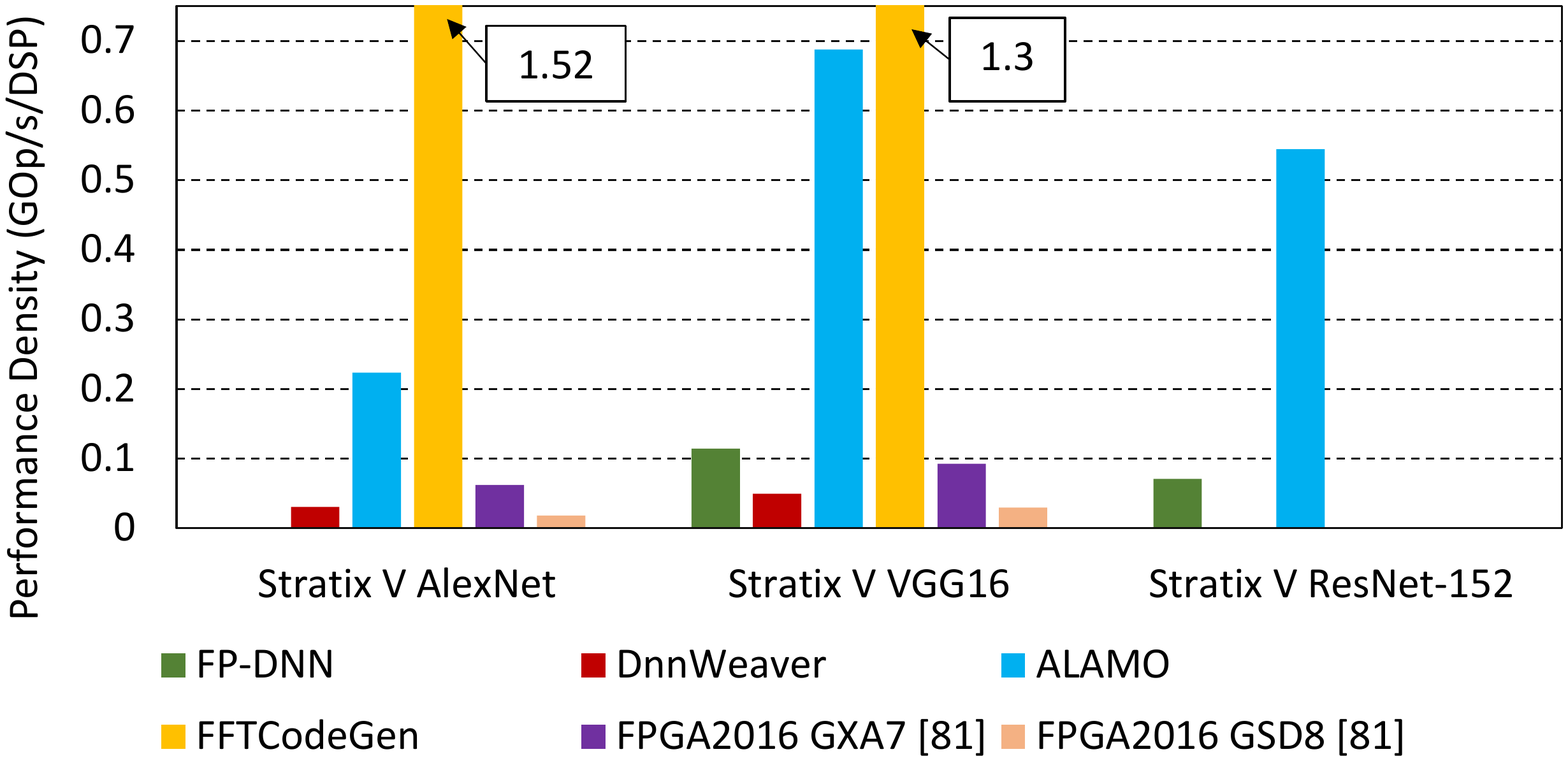}
    \vspace{-0.4cm}
    \caption{DSP-normalised comparison on mapping AlexNet, VGG16 and ResNet-152 on Stratix V}
    \label{fig:stratix_v_tools}
    \vspace{-0.5cm}
\end{figure*}

\textsc{DnnWeaver} and FP-DNN mapped VGG16 and VGG19 respectively on Stratix V GSD5 (Fig. \ref{fig:stratix_v_tools}). 
VGG19 has a larger workload compared to VGG16 by having three additional CONV layers. 
A larger number of CONV compared to FC layers can facilitate an accelerator's performance since CONV layers are computation bounded. Noting this difference in the VGG19 and VGG16 workloads, we use \mbox{FP-DNN's} performance on VGG19 as an indicator of its throughput on VGG16. In this respect, FP-DNN achieves a throughput of $364.36$ GOp/s and outperforms \textsc{DnnWeaver} by $2.31 \times$.

ALAMO maps AlexNet, VGG16 and ResNet-152 on Stratix V GXA7 which differs from the GSD5 device used by \textsc{DnnWeaver} and FP-DNN, {\color{black}with FFTCodeGen mapping AlexNet and VGG16 on the same device that is present on the Intel HARP platform}. Despite belonging to the same FPGA family, the two devices have been designed and optimised for applications with different characteristics. Stratix V GSD5 has been designed for algorithms with a large number of multiply-accumulate operations, while Stratix V GXA7 is optimised for high-bandwidth applications. In this respect, FP-DNN, \textsc{DnnWeaver}, ALAMO and FFTCodeGen are compared with respect to DSP-normalised throughput (Fig. \ref{fig:stratix_v_tools}). ALAMO demonstrates {\color{black}$6 \times$ and $13.89\times$ higher normalised throughput than FP-DNN and \textsc{DnnWeaver} on VGG16. Furthermore, ALAMO achieves $7.28\times$ higher normalised throughput than \textsc{DnnWeaver} for the mapping of AlexNet on Stratix V and $7.64\times$ higher normalised throughput than FP-DNN on ResNet-152 on Stratix V. Similarly, FFTCodeGen's VGG16 accelerator achieves $11.41\times$, $26.4\times$ and $1.9\times$ higher GOp/s/DSP than FP-DNN, \textsc{DnnWeaver} and ALAMO, with FFTCodeGen's AlexNet design demonstrating $49.66\times$ and $5.82\times$ higher GOp/s/DSP over \textsc{DnnWeaver} and ALAMO.} However, this metric does not capture the bandwidth difference of the two devices. Despite having fewer DSP blocks, the high bandwidth of Stratix V GXA7 enables ALAMO and FFTCodeGen to sustain a higher utilisation of their DSP resources. As a result, DSP-normalised throughput does not reflect the intrinsic strengths and weaknesses of the four designs since it does not take into account essential device-specific characteristics. {\color{black}Moreover, ALAMO is employing both DSPs and ALMs to implement its compute units, which enables the toolflow to reach higher performance than the majority of its DSP-based counterparts, while FFTCodeGen performs convolutions in the frequency domain with lower computational complexity and outperforms the competing toolflows.}

\begin{figure*}[t]
    \centering
    \includegraphics[trim={2cm 5cm 0cm 4.5cm},clip,width=0.9\textwidth]{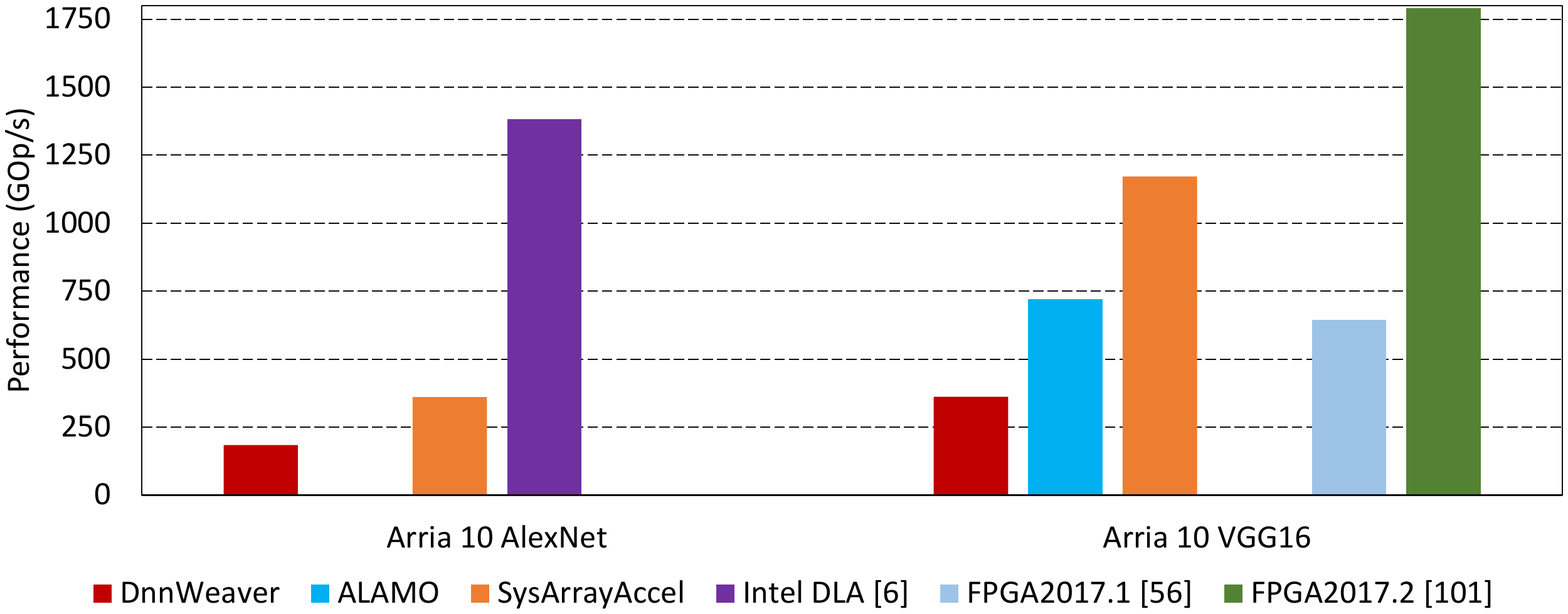}
    \vspace{-0.4cm}
    \caption{Comparison on mapping AlexNet and VGG16 on Arria 10}
	\label{fig:arria10_tools}
    \vspace{-0.5cm}
\end{figure*}

\textsc{DnnWeaver} and SysArrayAccel mapped AlexNet on Arria 10 GX115 and GT115 (Fig. \ref{fig:arria10_tools}). On AlexNet, SysArrayAccel achieves a throughput of 360.4 GOp/s and outperforms \textsc{DnnWeaver} by $1.95\times$. By targeting VGG16 on the same device, SysArrayAccel reaches $1.27\times$ (with FP precision) and $3.24\times$ (with 16-bit FXP precision) higher throughput than \textsc{DnnWeaver}, while outperforming ALAMO by $1.62\times$ (with 16-bit FXP precision) {\color{black}and with ALAMO overpassing by $1.56\times$ (with FP precision). AutoCodeGen and Caffeine are the only toolflows to target Virtex 7 VX690T and UltraScale KU060 respectively and therefore no meaningful comparison can be conducted with the rest of the toolflows.}

\textbf{Comparison between Toolflows Discussion.}
Among fpgaConvNet, \textsc{DnnWeaver}, DeepBurning {\color{black}and Snowflake}, fpgaConvNet's higher throughput comes potentially as a result of its SDF-based design methodology that allows for a finer exploration of the design space. On the other hand, \textsc{DnnWeaver}'s heuristic mapping and scheduling algorithm aims to optimally configure the templates of its accelerator, which leads to higher performance than DeepBurning, but still with less room for finer-grained customisation over a given CNN-FPGA pair than fpgaConvNet. With respect to latency, although DeepBurning {\color{black}and Snowflake} are designed to co-optimise latency and throughput by operating with a batch size of 1, fpgaConvNet's latency-driven methodology \cite{Venieris_2017b} is explicitly used for the generation of latency-optimised accelerators leading to a lower latency for AlexNet's feature extractor on both Zynq XC7Z020 and XC7Z045. {\color{black}Snowflake has been optimised from both an architectural and a compiler level to sustain close to peak utilisation of its compute resources and operate at the high clock frequency of 250 MHz. Despite utilising 3.5$\times$ fewer DSPs than DeepBurning, Snowflake demonstrates a 11\% higher throughput on AlexNet by reaching a computational efficiency of 94\%.} Overall, the analytical design space exploration methods of fpgaConvNet and \textsc{DnnWeaver}, which provide a finer optimisation of the hardware, proved to give a slight advantage over the brute-force mapping approach of DeepBurning.

By mapping VGG16 on Zynq XC7Z045, Angel-Eye outperforms the fpgaConvNet-generated design. By taking into account the clock frequency difference between the two designs, the clock frequency-normalised throughput ratio becomes $1\times$. Nevertheless, fpgaConvNet employs batch processing in order to achieve high throughput while Angel-Eye has been designed to co-optimise throughput and latency by operating with a batch size of 1. In this context, Angel-Eye achieves $1.52 \times$ lower raw latency
and $1.27 \times$ lower clock frequency-normalised latency. The latency gap between the two frameworks comes potentially from the fact that Angel-Eye's parameter space, as presented in Section \ref{param_space_section}, enables the tool to exploit the parallelism across both the input and output feature maps of each CONV layer, by partially unrolling both dimensions to minimise latency. On the other hand, fpgaConvNet's architecture tunably unrolls only the output feature maps and applies pipelining to the input feature maps which sets a higher bound on the latency.

{\color{black}For the mapping of LeNet-5 and CIFAR-10, \textsc{Haddoc2} employs 3-bit and 6-bit representations for both weights and feature maps. The toolflow implements all its compute units in logic instead of DSPs, and hence such low-precision operands enable the instantiation of a large number of units and the extraction of high throughput from the target device, without being limited by the available number of DSPs. On the other hand, fpgaConvNet employs 16-bit representation and implements its operators solely using DSPs. Consequently, \textsc{Haddoc2} outperforms fpgaConvNet in the particular set of benchmarks. Nevertheless, the requirement of \textsc{Haddoc2} for all weights to be stored on-chip and the mapping of all CNN computations to dedicated units limits the maximum size of CNNs that can be mapped to a particular device and, in this respect, \textsc{Haddoc2} can mainly support aggressively quantised networks.}

FP-DNN has combined OpenCL-based control circuitry with a hand-crafted RTL computation engine in order to overcome the limitations of OpenCL-generated compute units and exploit its advantages in the handling of control and interfacing with the host and the external memory. FP-DNN's scope is limited to data-centre setups which allows for server-specific assumptions and optimisations, such as PCIe-based communication. FP-DNN's highly optimised RTL Matrix Multiplication engine together with the sophisticated on-chip buffer allocation scheduling method have demonstrated the ability to target large-scale CNNs, such as VGG19 and ResNet-152 with a limitation on the supported FPGA setup. On the other hand, \textsc{DnnWeaver} demonstrates a wider scope, including both embedded and server-based FPGAs across different FPGA vendors and hence generality is a higher priority. {\color{black}ALAMO maps its compute units to both DSPs and logic in order to extract higher throughput from the target device and employs RTL-level design to perform low-level optimisations. These properties have enabled ALAMO to outperform FP-DNN and \textsc{DnnWeaver} on the same CNN models. On the other hand, FFTCodeGen performs convolutions in the frequency domain and achieves higher throughput than other toolflows on the same CNN-FPGA pairs by exploiting the lower computational complexity of this approach. Nevertheless, batch processing of the inputs is required by FFTCodeGen to amortise the overhead of converting between the space and frequency domains, which sets a limit to the lowest attainable latency of the framework.} Finally, SysArrayAccel employs its analytical performance and resource models in order to efficiently traverse the design space of systolic arrays and automatically configure tunable parameters at a finer grain than \textsc{DnnWeaver}. This leads to SysArrayAccel's higher performance on Arria 10.

\textbf{Comparison with Hand-Tuned FPGA Designs.}
ALAMO {\color{black}and FFTCodeGen} map AlexNet and VGG16 on Stratix V GXA7. We compare it with the design by Suda et al. \cite{Suda_2016} which targets the same CNN-FPGA pairs \mbox{(Fig. \ref{fig:stratix_v_tools})}. The design in \cite{Suda_2016} employs a formal optimisation formulation to configure an OpenCL-based accelerator that achieves $31.8$ GOp/s on AlexNet and $47.5$ GOp/s on VGG16. Compared to these designs, ALAMO achieves a speed-up of $3.6 \times$ on AlexNet and $7.41\times$ on VGG16, {\color{black} with FFTCodeGen outperforming by $38.07\times$ and $21.37\times$}. 

Despite the analytical design space exploration approach, Suda et al. use an off-the-shelf OpenCL matrix multiplication kernel for the CONV layers. ALAMO's RTL designs avoid the inefficiencies introduced by OpenCL and leave space for hardware optimisations at a lower level by mapping MACC units to both DSPs and logic. As a result, ALAMO trades off sophisticated design space exploration for highly optimised, RTL-based compute units to reach high performance. {\color{black}From a different perspective, the lower computational complexity of performing convolutions in the frequency domain and the generation of a highly optimised computation engine enable FFTCodeGen to outperform the OpenCL-based accelerator of \cite{Suda_2016}.}

Fig. \ref{fig:stratix_v_tools} also presents a comparison between FP-DNN and \textsc{DnnWeaver} with the design of Suda et al. on Stratix V GSD8. With the particular device belonging to the same class as Stratix V GSD5, DSP-normalised metrics provide a meaningful comparison. Both FP-DNN and \textsc{DnnWeaver} demonstrate higher performance than Suda et al. due to their highly optimised RTL designs.

Fig. \ref{fig:arria10_tools} presents a throughput comparison of the mapping of AlexNet and VGG16 on Arria 10 between toolflows and state-of-the-art, hand-crafted designs. For \textsc{DnnWeaver}'s and SysArrayAccel's AlexNet accelerators on Arria 10 GX115 and GT115, we compare with the state-of-the-art Deep Learning Accelerator (DLA) by Intel \cite{Aydonat_2017}. DLA achieves 1.382 TFLOp/s on AlexNet, with \textsc{DnnWeaver} and SysArrayAccel reaching $13.26 \%$ and $26 \%$ of DLA's performance. DLA employs a series of strategies to increase the DSP utilisation of Arria 10. These include the use of the Winograd transform to reduce the number of operations in convolutions and the design of a high-throughput 1D systolic array that operates at the high frequency of 303 MHz. Although DLA outperforms both \textsc{DnnWeaver} and SysArrayAccel with respect to AlexNet's mapping on Arria 10, it does not include an automated design flow and operates under the assumption that all intermediate feature maps can be cached on-chip. This is an assumption that is valid for networks of comparable size to AlexNet, but it does not hold for larger-scale models, such as VGG16 and ResNet-152.

For the VGG16 designs of \textsc{DnnWeaver} and SysArrayAccel on Arria 10, we compare with the accelerators presented in \cite{Ma_2017} and \cite{Zhang_2017}. \textsc{DnnWeaver} achieves $56.03 \%$ and $20.20 \%$ of the throughput of \cite{Ma_2017} and \cite{Zhang_2017} respectively. SysArrayAccel outperforms the throughput of \cite{Ma_2017} by $1.81 \times$ with a $1.78 \times$ improved latency and reaches $65.43 \%$ of the throughput of \cite{Zhang_2017} with $1.56 \times$ degraded latency. The higher performance of Zhang et al. \cite{Zhang_2017} comes due to the fact that the authors embedded RTL in OpenCL kernels in order to introduce register-level optimisations for particular networks, which enabled them to reach the high frequency of 385 MHz and sustain high utilisation of the FPGA's on-chip RAM. Despite achieving higher performance, such optimisations are hand-crafted and hence hard to exploit in other models in an automated manner.

{\color{black}For the AlexNet design of AutoCodeGen on Virtex 7 VX690T, we compare with the \mbox{565.94-GOp/s} accelerator presented in \cite{Li2017fpl}. AutoCodeGen achieves 222.1 GOp/s and reaches 39\% of the throughput of the highly optimised accelerator. The design in \cite{Li2017fpl} has achieved the current state-of-the-art performance of AlexNet on Virtex 7 VX690T, with manual optimisations for the particular CNN-FPGA pair. In this respect, despite achieving a lower raw throughput, AutoCodeGen is able to target a wider range of networks than \cite{Li2017fpl} by means of its automated flow.}

At the time of writing, Angel-Eye's mapping of VGG16 on Zynq XC7Z045 and fpgaConvNet's mappings of AlexNet on Zynq XC7Z020 and XC7Z045 and VGG16 on Zynq XC7020 are the state-of-the-art designs for the particular CNN-FPGA pairs. Similarly, Caffeine is currently the highest performing design to target Xilinx UltraScale KU060 and therefore no meaningful comparison can be made with hand-tuned designs at the moment.

\subsection{Discussion: Quality of Results}
The limitations of the ad-hoc benchmarking methodology of each toolflow do not allow us to draw conclusive and meaningful results on the comparative QoR of the generated accelerators. A uniform evaluation methodology that aims at surpassing the drawbacks of the current evaluation procedures for CNN-to-FPGA toolflows is proposed in Section \ref{unif_eval_section}. For the toolflows that target the same CNN-FPGA pairs, the following observations are made:

\textbf{1) Toolflows that generate highly optimised RTL-based designs tend to outperform their HLS counterparts.} This property can be observed in the comparison of ALAMO with the accelerator by Suda et al. \cite{Suda_2016}. Although Suda et al. employed a more sophisticated DSE method, ALAMO trades off a complex DSE to a detailed, RTL-level optimisation of its hardware design and outperforms by $3.6 \times$. A similar trend is observed between FP-DNN with \textsc{DnnWeaver}. Despite the fact that both toolflows generate RTL designs, FP-DNN focuses more on the low-level RTL optimisation of its computation engine and manages to achieve higher throughput, despite not performing the extensive design space exploration of \textsc{DnnWeaver}. The highest performing VGG16 accelerator on Arria 10 by Zhang et al. \cite{Zhang_2017} provides additional evidence. The design in \cite{Zhang_2017} embeds custom RTL-level optimisations in OpenCL kernels to boost the performance and avoid the current limitations of the OpenCL programming model. In this way, \cite{Zhang_2017} achieves higher performance than the CNN-to-FPGA toolflows. In spite of this advantage, the manual, hand-crafted RTL design that is required to achieve this level of performance prohibits the automation that is essential for a toolflow and therefore a trade-off between RTL performance and HLS productivity is necessary.

\textbf{2) Design space exploration methods that allow for finer customisation tend to offer an advantage in terms of \mbox{Quality} of Results.} fpgaConvNet's analytical methodology outperformed \textsc{DnnWeaver}'s slightly more restricted design space, which in turn outperformed DeepBurning's heuristic mapping. Moreover, SysArrayAccel's detailed design space exploration method enabled the traversal of a larger design space than \textsc{DnnWeaver}, which led to higher performing designs. {\color{black}Similarly, the CaP technique (Section \ref{hw_arch}) introduced by FFTCodeGen added another level of customisation and enabled to full exploitation of the FFT-based convolution by sustaining a high utilisation of the generated accelerator across CONV layers of different sizes.}

\textbf{3) Single computation engine architectures tend to reach high performance on CNNs with a uniform structure.} This property can be observed in the case of the increase in the throughput of \textsc{DnnWeaver} and SysArrayAccel when mapping VGG16 compared to AlexNet. The single computation engine of both toolflows manages to sustain a very high utilisation across the layers of VGG16 due to the uniform kernel size of the CONV layers and the power of 2 number of input and output feature maps after the first CONV layer. In contrast, the variable kernel sizes of AlexNet, including $11 \times 11$, $5 \times 5$ and $3 \times 3$, lead to an underutilisation of the shared computation engine. Moreover, Angel-Eye's mapping of VGG16 on Zynq XC7Z045 is also benefited by the uniformity of VGG16 and reaches the highest reported raw performance for VGG16 on the particular device. {\color{black}Nevertheless, FFTCodeGen is not affected by the irregularity in the kernel sizes of AlexNet due to its tiled FFT-based algorithm and hence its throughput is not deteriorated.}

\subsection{Discussion: Suitability for Deep Learning Application Challenges}
{\color{black}CNNs have been successfully employed in a variety of problem domains, including video surveillance \cite{Shao2017crowd_understanding}, healthcare \cite{Esteva2017} and autonomous transportation \cite{Chen2015deepdrive}. Depending on the nature of the domain, CNNs have to be deployed on processing platforms with different constraints and compute capabilities, spanning from server-grade setups in a data centre \cite{Caulfield2016} to low-power devices on the edge \cite{Smolyanskiy_2017}. Moreover, the variability of applications requires from the CNN implementations to comply with diverse performance requirements, from the high-throughput needs of large-scale cloud-based services to the critical low-latency requirements of autonomous drones and cars, with low power consumption standing as a ubiquitous requirement. In this context, the different design approaches of the existing CNN-to-FPGA toolflows determine their suitability to particular use cases in the deep learning application landscape.}

\begin{figure}[t] 
    \vspace{-0.5cm}
	\centering
	\includegraphics[trim={3cm 3.5cm 3.5cm 2.2cm},clip,width=1\linewidth]{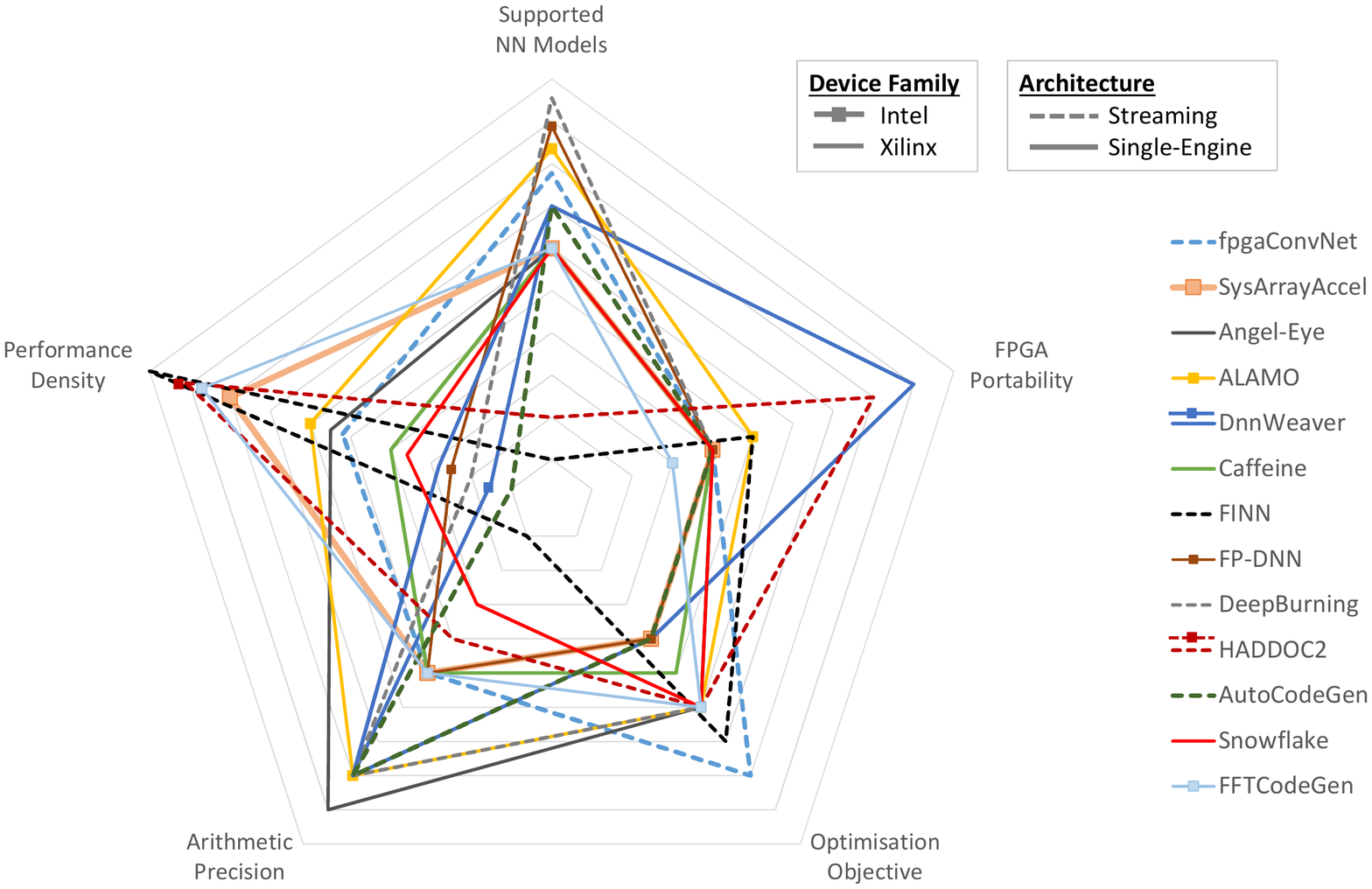}
	\caption{Overview of toolflow characteristics}
	\label{fig:spider}
	\vspace{-0.5cm}
\end{figure}

Table \ref{frameworks_table} summarises the features of each toolflow and the chart in Fig. \ref{fig:spider} depicts the effect of each toolflow's strategic design decisions on a number of aspects. \textsc{Finn} trades-off very high throughput and low latency by restricting its focus on the fine niche of binarised neural networks. The toolflow's Vivado HLS-based accelerators can target only Xilinx devices, but with enough infrastructure to target both embedded SoCs and standalone devices. {\color{black}These properties make \textsc{Finn}-generated designs to be lightweight and applicable to both throughput-driven and latency-sensitive applications, that also exhibit high error tolerance, due to the potential impact of binarisation on the accuracy.}

\begin{table*}[t]
\centering
\vspace{-0.0cm}
\caption{Summary of Toolflow Characteristics}
\vspace{-0.25cm}
\label{frameworks_table}
\resizebox{\linewidth}{!}{%
\setlength{\tabcolsep}{2pt}

	\begin{tabular}{l l l l l l l}
		\toprule
		\multirow{1}{*}{Framework Name} & \multicolumn{1}{l}{\multirow{1}{*}{Interface}} & NN models & Devices & Architecture  & Precision$^*$ & DSE  \\ 
		\midrule 
		
		fpgaConvNet \cite{Venieris_2016}\cite{Venieris_2017}\cite{Venieris_2017b}\cite{Venieris_2017c}      & \begin{tabular}[t]{@{}l@{}} Caffe \& Torch \end{tabular}                                            & CNN,Res,Incep,Dense  & Xilinx SoC  & Streaming & FXP (Uniform) \& FP  & Global Optimiser (Simulated Annealing) \\
		
		DeepBurning \cite{Wang_2016}  & Caffe  & CNN,RNN,DNN   & Xilinx SoC & Streaming & FXP (Dynamic) & Heuristic \\
		
		Angel-Eye \cite{Qiu_2016}\cite{Guo_2016}\cite{Guo_2018}   & \begin{tabular}[t]{@{}l@{}} Caffe \\ 
		\end{tabular}  & CNN,DNN   & Xilinx SoC  & Single-Engine & FXP (Dynamic) & Heuristic with Analytical Model \\
		
		ALAMO \cite{Yufei_Ma_2016}\cite{Ma_2017}\cite{Ma_2017b}\cite{Yufei_Ma_2017}\cite{Ma_2018} & \begin{tabular}[t]{@{}l@{}} Caffe 
		\end{tabular} & CNN,DNN  & Intel SoC \& Standalone & Single-Engine & FXP (Dynamic) & Heuristic \\
		
		{\color{black} \textsc{Haddoc2} \cite{Abdelouahab_2016}\cite{Abdelouahab_2017}} & {\color{black}Caffe} & {\color{black}CNN,DNN} & {\color{black}Xilinx \& Intel Standalone} & {\color{black}Streaming} & {\color{black}FXP (Uniform)} & {\color{black}Deterministic} \\
		
		\textsc{DnnWeaver} \cite{Sharma2016DnnweaverFH}\cite{Sharma_2016}   & Caffe & CNN,DNN  & Xilinx \& Intel  & Single-Engine & FXP (Dynamic) & Custom Search Algorithm \\ 
		
		Caffeine \cite{Zhang_2016} & Caffe & CNN,DNN & Xilinx Standalone & Single-Engine & FXP (Uniform) \& FP & Exhaustive over Roofline Model \\
		
		{\color{black}AutoCodeGen \cite{Zhiqiang_Liu_2016}} & {\color{black} Proprietary Input} & {\color{black}CNN,DNN} & {\color{black} Xilinx Standalone} & {\color{black}Streaming} & {\color{black}FXP (Dynamic)} & {\color{black}Heuristic with Analytical Model} \\
		
		\textsc{Finn} \cite{Umuroglu_2017}\cite{Fraser_2017} & Theano & BNN & Xilinx SoC \& Standalone & Streaming & Binary & Heuristic \\
		
		FP-DNN \cite{Guan2017} & TensorFlow &  CNN,RNN,DNN,Res & Intel Standalone  & Single-Engine & FXP (Uniform) \& FP  & Algorithmic \\
		
		{\color{black}Snowflake \cite{Gokhale2017}\cite{Chang_2017}} & {\color{black} Torch} & {\color{black}CNN,Res,Incep} & {\color{black}Xilinx SoC} & {\color{black}Single-Engine} & {\color{black}16-bit FXP (Uniform)} & {\color{black}Heuristic} \\
		
		SysArrayAccel \cite{Wei2017} & C Program & CNN,DNN & Intel Standalone & Single-Engine & FXP (Uniform) \& FP  & Exhaustive over Analytical Model \\
		
		{\color{black}FFTCodeGen \cite{fft2017fpga}\cite{fftcodegen2017rpt}\cite{fft2017reconfig}\cite{fft2018fpga}} & {\color{black}Proprietary Input} & {\color{black}CNN,DNN} & {\color{black}Intel HARP} & {\color{black}Single-Engine} & {\color{black}FXP (Uniform) \& FP}  & {\color{black}Roofline and Analytical Models} \\
		
	\multicolumn{3}{l}{* FXP: Fixed-Point, FP: Floating-Point.} \\
		\bottomrule
	\end{tabular}
}
\vspace{-0.65cm}
\end{table*}

\textsc{DnnWeaver} places FPGA compatibility and portability as a priority and bases its internal design on device-independent, RTL-based templates. The toolflow's infrastructure and template-level parametrisation allow for variable precision along the CNN layers and has demonstrated the widest support for SoCs, standalone and server-grade FPGAs from different vendors. Its application scope is restricted to high-throughput applications with large batch sizes, without special consideration for low-latency requirements, which restricts the toolflow's supported optimisation objectives.

FP-DNN places emphasis on the low-level optimisation of a computation engine that would support different types of NN models. The toolflow restricts its scope to cloud-based environments with Intel FPGAs and is optimised for the high-throughput workloads of data centres. The toolflow adopts uniform quantisation across the CNN layers as specified by the user.

Caffeine and SysArrayAccel concentrated on the optimisation of systolic array structures for high-throughput CNNs. The two toolflows are restricted to Xilinx and Intel FPGAs due to their use of Vivado HLS and OpenCL respectively. SysArrayAccel draws from the lessons learned from Caffeine's design and its finer design space exploration method tends to yield higher performance. However, the different target devices of the two toolflows do not allow for a meaningful performance comparison.

ALAMO is designed to combine the high throughput and low latency of RTL designs with high precision flexibility across layers. Nevertheless, precision quantisation is performed manually and is not part of the automated flow. The toolflow uses Intel's off-the-shelf IPs, including the NIOS soft processor and the scatter-gather DMA block, and hence the generated designs are restricted and tailored for Intel FPGAs, which affects its portability across vendors.

Angel-Eye bases its competitive advantage on its automatic dynamic quantisation scheme. The selected parameter space allows for the unrolling of both the input and output feature maps and hence latency and throughput are co-optimised.

fpgaConvNet prioritises the support of various optimisation objectives based on the application-level performance needs in order to target diverse workloads. In this context, distinct methodologies are used for high-throughput, low-latency or multiobjective applications. Similarly to Caffeine and Angel-Eye, the use of Vivado HLS currently restricts fpgaConvNet to Xilinx devices.

DeepBurning's design principle entails modularity and support of a wide range of NN models. In this respect, the toolflow's RTL building blocks can target various types of NNs, including the emerging RNNs and LSTMs, and offer flexibility with respect to precision quantisation across the layers. By design, the generated accelerators are optimised to operate with a batch size of 1 and hence their optimisation objectives are simultaneously high throughput and low latency. {\color{black}Similarly, AutoCodeGen also places focus on the modular design of RTL hardware blocks for high throughput, with a more restricted scope than DeepBurning by supporting only CNN models and with the addition of high-level performance and resource modelling.}

{\color{black}\textsc{Haddoc2} follows a direct mapping approach, where all layers and neurons in a network are mapped deterministically to dedicated hardware resources. This approach enables achieving both high throughput and low latency in the cases where all weights of the target CNN can be accommodated by the on-chip memory resources and enough logic is available to map all operators. This assumption holds in the case of small-scale networks, such as LeNet-5 and CIFAR-10, and aggressively quantised models, such as the BNNs targeted by \textsc{Finn}, but \textsc{Haddoc2} cannot currently handle the state-of-the-art large-scale models, due to the lack of tunable time-sharing mechanisms.}

{\color{black}Snowflake's design principle places programmability and high utilisation of the computational resources at the forefront. In this respect, both Snowflake's architecture and compiler are tailored to removing inefficiencies and extracting close to peak performance from the allocated resources. Overall, Snowflake favours programmability over hardware specialisation, by employing a fixed hardware design and customising with respect to the target model only at the compiler level.}

{\color{black}Finally, FFTCodeGen addresses CNN acceleration from both an algorithmic and an architectural level. In contrast to the rest of the toolflows, convolutions are performed in the frequency domain with a significantly lower computational complexity. Moreover, the free parameters of the algorithm and the architecture enable the generated compute engine to sustain high throughput across convolutional layers of different sizes and fully exploit the computational complexity gains. Furthermore, the use of the powerful, server-grade CPU of the target Intel HARP platform alleviates the complexities of mapping the memory-bounded fully-connected layers to hardware and further contributes to FFTCodeGen's throughput gains, making it suitable for throughput-driven cloud-based applications.}

\subsection{Other Related Work}
{\color{black}
Apart from the presented toolflows, several FPGA-based designs for CNNs have been proposed by the FPGA community. These include highly optimised, hand-tuned accelerators for particular CNN-FPGA pairs in RTL \cite{Farabet_2010}\cite{Dundar_2017}\cite{Li2017fpl}, HLS \cite{Aydonat_2017}\cite{Kim2017legupdnn} and mixed RTL-HLS \cite{Zhang_2017}, together with designs that focus on optimising the external memory bandwidth utilisation \cite{Alwani2016micro}\cite{Yongming2017}. A number of existing works lie close to the presented CNN-to-FPGA toolflows, but lack essential components that would form a complete automated flow. These include \cite{Motamedi_2016}\cite{Suda_2016}\cite{Motamedi2017placid}\cite{DiCecco2016fpt}, with \cite{Motamedi_2016}\cite{Suda_2016}\cite{Motamedi2017placid} focusing on the design space exploration task and \cite{DiCecco2016fpt} presenting an FPGA back end to Caffe, for the execution of $3 \times 3$ convolutional layers by means of the Winograd transform.
}

\section{The Future of CNN-to-FPGA Toolflows}

\subsection{Towards a Uniform Evaluation Methodology}
\label{unif_eval_section}

The existing FPGA tooflows have employed ad-hoc evaluation methodologies, by targeting different CNN models and reporting the achieved performance in a non-uniform manner. A uniform evaluation methodology is proposed here in order to enable the thorough and comparative evaluation of CNN-to-FPGA toolflows. The proposed methodology comprises a benchmark suite and guidelines for evaluation metrics.

\textbf{Benchmark Suite.}\footnote{The proposed benchmark suite of representative CNNs can be found in: http://www.imperial.ac.uk/intelligent-digital-systems/cnn-benchmark-suite/} 
A comprehensive benchmark suite should include CNNs that are widely used and whose accuracy has been extensively studied by the deep learning community. Each CNN should pose a unique hardware mapping challenge and stress the FPGA toolflow from a different aspect. The main challenges to be addressed include CNNs that are (1) computation bounded, (2) off-chip memory bandwidth bounded, (3) on-chip memory capacity bounded, (4) with high layer dependency and (5) irregular and sparse layer connectivity that challenges scheduling.

\begin{table}[t]
    \vspace{-0.2cm}
	\centering
	\caption{Benchmark Suite: CNN Models and their Computational Challenges}
	\vspace{-0.25cm}
	\label{cnn_models_table}
	\resizebox{0.75\linewidth}{!}{
		\begin{tabular}{l r l c l l c c c}
			\toprule
			Model Name & & Year & Depth & Design Principles & Challenges \\
			\midrule
			
			AlexNet & \cite{Krizhevsky2012} & 2012 & 8 & \begin{tabular}[t]{@{}l@{}} 1) Increased depth \\ 2) Increased layer width \end{tabular} & \begin{tabular}[t]{@{}l@{}} 1) Non-uniform filter sizes \\ 2) Grouped convolutions \end{tabular} \\
			
			ZFNet & \cite{Zeiler_2014} & 2013 & 8 & \begin{tabular}[t]{@{}l@{}} 1) Wider layers \end{tabular} & \begin{tabular}[t]{@{}l@{}} 1) Computational load \\ 2) Memory footprint \end{tabular} \\
			
			VGG16 & \cite{Simonyan14c} & 2014 & 16 & \begin{tabular}[t]{@{}l@{}} 1) Increased layer depth \\ 2) Uniform filter size \end{tabular} & \begin{tabular}[t]{@{}l@{}} 1) Computational load \\ 2) Memory footprint \\ 
			\end{tabular} \\
			
			GoogLeNet & \cite{Szegedy2014} & 2014 & 22 & 1) Inception module & \begin{tabular}[t]{@{}l@{}} 1) Irregular computations \\ 2) Irregular layer connectivity \end{tabular} \\
			
			ResNet-152 & \cite{He_2016} & 2015 & 152 & \begin{tabular}[t]{@{}l@{}} 1) Residual block \end{tabular} & \begin{tabular}[t]{@{}l@{}} 1) Irregular computations \\ 2) Irregular layer connectivity \end{tabular} \\ 
			
			
			Inception-v4 & \cite{Szegedy_2016} & 2016 & 72 & 1) Residual Inception block & \begin{tabular}[t]{@{}l@{}} 1) Irregular computations \\ 2) Irregular layer connectivity \end{tabular} \\
			
			DenseNet-161 & \cite{huang2017densely} & 2017 & 161 & \begin{tabular}[t]{@{}l@{}} 1) Dense block 
			\end{tabular} & \begin{tabular}[t]{@{}l@{}} 1) Irregular computations \\ 2) Irregular layer connectivity \end{tabular} \\ 
			
			\bottomrule
		\end{tabular}%
	}
\vspace{-0.5cm}
\end{table}

To this end, we propose a benchmark suite with the following CNN models (Table \ref{cnn_models_table}): AlexNet, ZFNet, VGG16, GoogLeNet, ResNet-152, Inception-v4 and {\color{black}DenseNet-161}. AlexNet was the winner of the 2012 ILSVRC competition and its pretrained feature extractor is widely used as a starting point for new applications \cite{Redmon_2015}, making it a fundamental CNN model in the deep learning community. AlexNet's mapping challenge lies in the non-uniform filter and stride sizes across its convolutional layers, including $11 \times 11$, $5 \times 5$ and $3 \times 3$ kernels which can stress the utilisation of convolution engines and assess a toolflow's mapping capabilities. As an example, the Angel-Eye computation engine is currently tailored to $3 \times 3$ kernels and optimised for the VGG16 network which has a uniform filter size. In this case, AlexNet could evaluate Angel-Eye's efficiency of mapping CONV layers that do not have $3 \times 3$ kernels. Moreover, fpgaConvNet derives a single streaming architecture for low-latency designs. The mapping of AlexNet would examine the quality of the derived streaming architecture of fpgaConvNet. In a similar manner, ZFNet also has non-uniform filter sizes, consisting of $7 \times 7$, $5 \times 5$ and $3 \times 3$ kernels, and wider CONV layers than AlexNet, while being used as a starting template for novel applications \cite{Ren_2017}. Both networks can lead to the exposure of finer strengths and weaknesses and indicate potential room for improvements in the toolflows.

VGG16 is a substantially deep model with high computation and memory requirements and constitutes one of the most widely employed pretrained models for new applications \cite{Badrinarayanan_2017}. Due to its large computational load, number of weights and layers, it is proposed as a representative neural network that poses challenges (1), (2), (3) and (4). Because of these challenges, the majority of existing tools have already evaluated their design flows on VGG16.

To challenge the CNN-to-FPGA frameworks with irregular and sparse computations, the benchmark suite includes GoogLeNet, ResNet-152, Inception-v4 {\color{black}and DenseNet-161}. GoogLeNet introduced the Inception module that made the CNN topology more complex than conventional CNNs by breaking the uniform layer connectivity. ResNet-152 introduced a residual block that allows for a forward connection that bypasses intermediate layers and enabled the construction of a 152-layered network. Inception-v4 increases the CNN complexity by combining both concepts from GoogLeNet and ResNet-152 in order to achieve higher performance. Despite the higher performance of Inception-v4, pretrained versions of GoogLeNet and ResNet-152 are still widely used. {\color{black}Finally, DenseNet-161 presented a dense block that enables the output of each layer to be directly connected to the input of every following layer.} This type of networks would provide a thorough evaluation of the CNN-to-FPGA toolflows. FP-DNN, ALAMO and fpgaConvNet have already demonstrated their performance when targeting ResNet-152, with Snowflake targeting ResNet-50 and GoogLeNet. Since FP-DNN consists of a single Matrix Multiplication engine, the ResNet-152 mapping reduced to the problem of scheduling the different layers given their irregular connectivity. {\color{black}In a similar manner, Snowflake's compiler breaks down residual blocks and Inception modules into MACC operations and schedules them over Snowflake's accelerator. On the other hand, ALAMO and fpgaConvNet followed a different approach by enhancing their architectures with specialised blocks to handle irregular networks. ALAMO designed and integrated a dedicated elementwise addition block in its computation engine to support the residual connections of ResNets. fpgaConvNet has introduced three specialised streaming hardware blocks, tailored for the Inception module and the residual and dense blocks.} Overall, toolflows that generate streaming architectures would have to cope with not breaking the streaming principle of operation and demonstrate how their mapping and scheduling methods can compare with the more flexible, single computation engine designs. 

\textbf{Evaluation Metrics.}
\label{eval_metrics_sec}
Evaluation metrics aim to characterise the quality of a toolflow's generated results and highlight the various strengths and weaknesses. These metrics should include essential attributes such as performance including throughput and latency, resource consumption, power efficiency and application-level accuracy. Reporting all these criteria play an important role in determining the strategic trade-offs made by a toolflow.

In terms of \textit{performance}, the most commonly reported metrics are currently affected by two limitations: (1) normalised quantities such as GOp/s/Logic and GOp/s/DSP attempt to indicate the quality of the generated design solely as a measure of computational resource utilisation. This approach does not capture the available bandwidth and capacity of the off- and on-chip memory which can have a decisive effect on performance; (2) normalising with a resource that is FPGA family-specific, such as Xilinx's LUTs or Intel's ALMs, does not enable the fair comparison across different vendors and across FPGA families from the same vendor. As a characteristic example, we point to the Stratix V devices targeted by FP-DNN and ALAMO, namely Stratix V GSD5 and GXA7. Despite the fact that both devices belong to the Stratix V family, GSD5 belongs to the GS FPGAs which are optimised for DSP-focused applications with an abundance of MACC operations and hence contains $6.2 \times$ more DSPs, smaller on-chip memory and fewer ALMs, while GXA7 belongs to the GX FPGAs which are optimised for high-bandwidth applications and hence offers a higher bandwidth interface to the off-chip memory, $1.14 \times$ larger on-chip memory and $1.36 \times$ more ALMs. As a result, a single resource-normalised metric such as DSP- or logic-normalised performance is not able to capture the quality of the generated hardware across devices that are optimised for different application domains.

\textit{Throughput} is the primary performance metric of interest in throughput-driven applications such as high-throughput image recognition and large-scale, multi-user analytics services over large amounts of data. Throughput is measured in GOp/s and is often achieved by processing large batches of inputs. \textit{Latency}, or \textit{response time}, becomes the primary critical factor in latency-sensitive applications, such as self-driving cars and autonomous systems, but also in particular real-time cloud-based services. Measured in seconds, latency is the time between when an input enters the computing system and when the output is produced. In such scenarios, batch processing adds a prohibitive latency overhead and is often not an option. Different toolflows choose to either optimise for one of the two metrics, co-optimise them simultaneously by using a batch size of 1 or selectively optimise for one of the two based on the application's performance requirements.

\textit{Resource consumption} is an indicator of the efficiency of the utilisation of the available resources on the target platform by the designs generated by a toolflow, including the DSPs, on-chip RAM, logic and FFs. \textit{Application-level accuracy} is a crucial metric when approximation techniques are employed by a toolflow for the efficient mapping of CNNs. Such techniques may include precision optimisation, such as the dynamic precision quantisation scheme by Angel-Eye, or lossy compression methods, such as the SVD-based compression applied on the weights of the FC layers by Angel-Eye, and can have an impact on the application-level accuracy of the CNN. Potential performance-accuracy trade-offs have to be quantified and reported in terms of accuracy degradation.

To measure the quality of a CNN-to-FPGA toolflow, we propose the following methodology. Throughput in GOp/s with explicitly specified GOp/network, amount of weights and batch sizes, and latency in seconds/input with batch size of 1, in order to present the throughput-latency relationship, should be included in the evaluation reports. Resource-normalised metrics are meaningful when comparing designs that target devices from the same FPGA family optimised for the same application domain. In this scenario, performance normalised with respect to logic and DSPs would allow the comparison of hardware designs for the same network on FPGAs of the same family. Power-normalised throughput and latency should also be reported for comparison with other parallel architectures such as CPUs, GPUs and DSPs. Since resource-normalised performance does not capture the effect of off- and on-chip memory bandwidth and capacity despite being critical for achieving high performance, target FPGA platform details should be included that explicitly indicate the off- and on-chip memory specifications. All measurements should be made for various CNNs, with emphasis on the proposed benchmark suite of the previous section, in order to demonstrate the quality of results subject to the different mapping challenges posed by each benchmark model.

\vspace{-0.2cm}
\subsection{Objectives of a CNN-to-FPGA Toolflow}

Recent developments in deep learning have led to new challenges in the deployment of CNNs. This section presents a set of objectives for CNN-to-FPGA toolflows based on recent research trends.

\begin{figure}[t] 
 	\vspace{-0.2cm}
	\centering
	\includegraphics[trim={0cm 10.25cm 0cm 10.75cm},clip,width=0.8\linewidth]{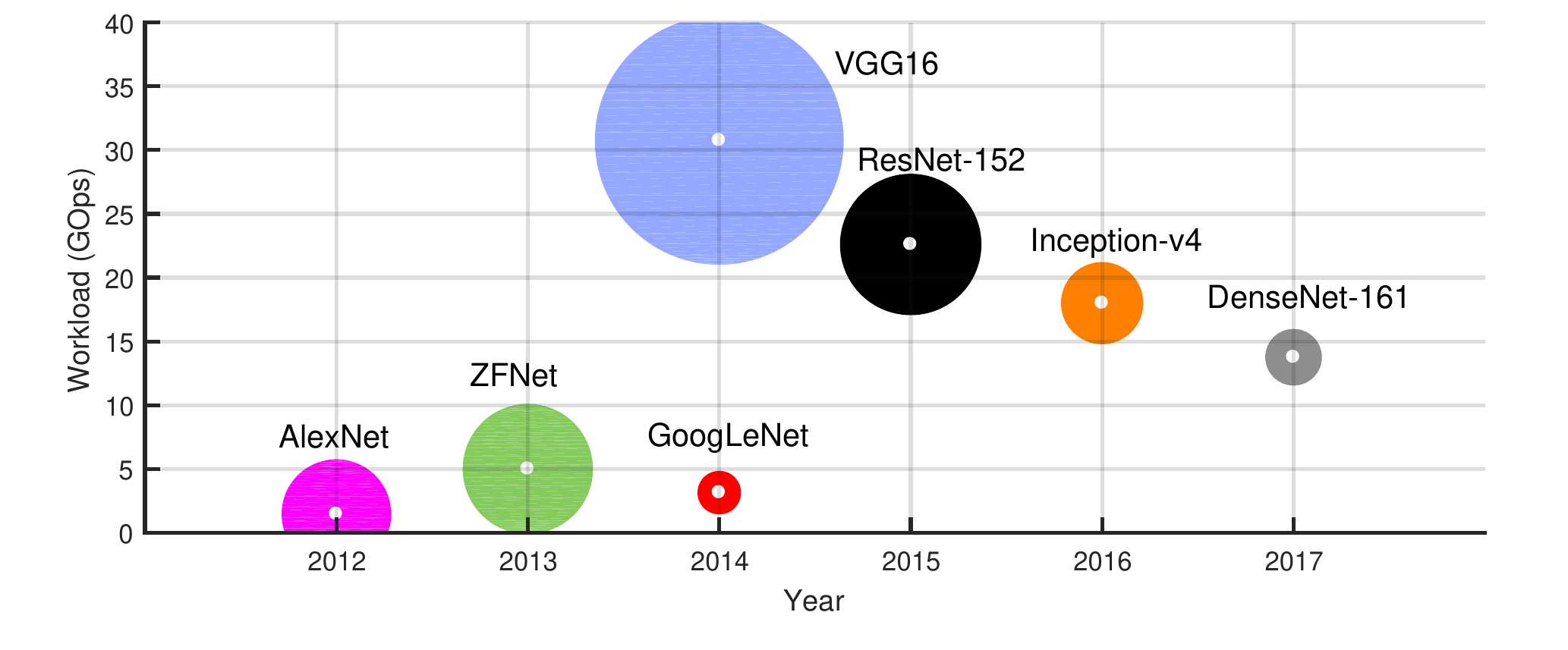}
	\vspace{-0.5cm}
	\caption{CNN computation and memory evolution (circle size is relative to the amount of weights)}
	\label{fig:cnn_schar_space}
	\vspace{-0.5cm}
\end{figure}

\textbf{\textit{Objective 1}. Targeting next-generation CNN models.}
Since AlexNet's win in the 2012 ILSVRC competition, a number of CNN models \cite{Zeiler_2014}\cite{Simonyan14c}\cite{Szegedy2014}\cite{He_2016}\cite{Szegedy_2016}\cite{huang2017densely} paved the way for the state-of-the-art accuracy in visual tasks. Typically, improvements in accuracy have been achieved at the expense of increased complexity in the structure of the CNN. In Table \ref{cnn_models_table}, which lists a number of representative models together with the source of challenge in their design (also visualised in \mbox{Fig. \ref{fig:cnn_schar_space}}), three main trends are identified in CNN design: (1) the increase in the depth of the models, from the 8-layer AlexNet up to the 152-layer ResNet-152 {\color{black}and the 161-layer DenseNet-161}, (2) the increased inference workload, with an increase of $20 \times$ from AlexNet to VGG16 in GOps/input and (3) the introduction of novel compound components. With respect to (3), networks such as GoogLeNet, ResNet-152, Inception-v4 and {\color{black}DenseNet-161} enhanced the CNN layers with the introduction of complex blocks, as indicated in Table \ref{cnn_models_table}. This type of complex blocks break the uniform connectivity and computation pattern of conventional CNNs with irregular layer connections and challenge the automation of their mapping to hardware. {\color{black}Currently, ALAMO, Snowflake, FP-DNN and fpgaConvNet provide optimised support for residual blocks in networks. Moreover, Snowflake and fpgaConvNet also target networks that use Inception modules, such as GoogLeNet. Finally, fpgaConvNet supports dense blocks by means of a specialised hardware building block, tailored to dense block structures. Nevertheless, with such compound components becoming mainstream in the deep learning literature, CNN-to-FPGA toolflows ought to investigate further the optimisation opportunities of their mapping to optimised hardware.}

Beyond the spatial pattern recognition abilities of CNNs, Recurrent Neural Networks (RNNs) enhance NNs with the ability to learn data sequences by retaining memory \cite{mikolov2010recurrent}. While this broadens the application field of neural networks, additional recurrent connections between layers introduce further challenges in the parallelisation of computations. RNN models, with the prominence of LSTM networks \cite{lstm1997}, demonstrate state-of-the art accuracy in applications that require capturing long-range dependencies by processing information from past inputs \cite{Vinyals_2017}. Thus, designing optimised hardware units that support network architectures with such recurrence in connections between layers is becoming an increasingly important feature for FPGA toolflows. 

In contrast to the computation-bounded CNNs, RNNs and LSTMs comprise inherently memory-bounded workloads due to the large number of matrix-vector multiplications. This property necessitates a different design approach for their optimised mapping to hardware. At the moment, DeepBurning and FP-DNN offer support for RNNs, with FP-DNN also targeting LSTMs. In the same direction, the authors of Angel-Eye together with DeePhi\footnote{http://www.deephi.com/} have proposed an LSTM accelerator \cite{Han_2017}, but it has not been integrated into Angel-Eye. {\color{black}The FPGA community has further proposed designs targeting from high-throughput data-centre services \cite{Nurvitadhi_2016} to latency-critical embedded setups \cite{approxlstm2018arc}. Moreover, industrial companies such as Google \cite{Jouppi2017} and Microsoft\footnote{https://www.microsoft.com/en-us/research/blog/microsoft-unveils-project-brainwave/} report that a large fraction of their data-centre workloads are LSTM-based and have focused their efforts on optimising the execution of LSTMs with customised ASIC and FPGA designs respectively. In this context, along with the automated mapping of CNNs, end-to-end frameworks that would focus on the high-performance deployment of RNNs and LSTMs on FPGAs emerges as an essential objective.}

\textbf{\textit{Objective 2}. Support of compressed and sparse CNNs.}
Recent work from the deep learning community has proposed techniques of reducing the inference time of CNNs, by exploiting the redundancy across its trainable parameters. The existing approaches can be divided into \mbox{(1) post-training} and (2) training-time methods. Post-training methods assume fully trained CNNs and add a preprocessing step prior to deployment. Works such as \cite{Denton_2014}\cite{Jaderberg14b} focus on decreasing the computational cost of the computation-bounded CONV layers by means of the low-rank decomposition of filters. On the other hand, works such as \cite{SVD_2013}\cite{HanMD15} focus on minimising the excessive memory footprint of the memory-bounded FC layers, by projecting the weights matrix to a lower-dimensional space. Training-time methods attempt to create sparse CNNs by means of pruning or sparsity regularisation techniques during the training phase \cite{NIPS2015_5784}\cite{Baoyuan_Liu_2015}. Although the sparsification of CNNs can reduce the theoretical computational and memory costs, the elimination of weights and connections between layers breaks the uniformity of computation and memory accesses, and hence requires a rethinking of the hardware mapping. At the moment, a few ASIC designs have been proposed to tackle the challenges of compressed and sparse networks \cite{Han2016}\cite{cambriconx_2016}\cite{Parashar_2017}. In this context, there is an emerging need for the CNN-to-FPGA tools to support compressed and sparse networks in order to offer competitive high-performance, low-power alternatives to the existing CPU, GPU and DSP platforms.

\textbf{\textit{Objective 3}. Support of low-precision and extremely low-precision CNNs.}
The robustness of CNNs to low-precision quantisation of weights and feature maps at the inference stage has been widely studied \cite{Holi_1993}\cite{Gupta_2015}\cite{Gysel_2016}\cite{Hashemi_2017}\cite{Zhou_17}. At the moment, the majority of CNN-to-FPGA toolflows support either uniform or {\color{black}dynamic} quantisation across layers, depending on whether the wordlengths {\color{black}and scaling} at each layer are the same. Angel-Eye, ALAMO, \textsc{DnnWeaver}, DeepBurning and AutoCodeGen support dynamic quantisation, {\color{black}with a fixed, uniform wordlength and different scaling across layers}. The focus on quantisation has been taken a step further by Angel-Eye which employs {\color{black}an automated quantisation method to automatically determine the scaling for each layer of the target network, given a fixed wordlength}. {\color{black}Nevertheless, existing works have been investigating more irregular quantisation schemes, with the ASIC designs of \cite{Judd2016stripes}\cite{Albericio2017pragmatic} 
varying both the wordlength and scaling of each layer of the network and mapping the variable-wordlength computations on optimised bit-serial arithmetic units, and Intel Nervana proposing a custom floating-point variant format \cite{flexpoint2017nips}.} The robustness of CNNs to quantisation offers CNN-to-FPGA toolflows the opportunity to explore and integrate automatic quantisation methodologies as part of their design flows. Adding precision quantisation as a design dimension can offer more room for customisation in the architectural design space, provide a closer coupling of network and hardware design and offer more room for improvement over CPU, GPU and DSP counterparts which cannot benefit from this type of fine-grained data representation optimisations.

In the same context, Ternary \cite{zhu2016trained}\cite{Alemdar2017ijcnn} and Binary \cite{NIPS2016_bnns}\cite{rastegariECCV16} CNNs form extreme -but widely studied- cases of low-precision CNNs. It has been demonstrated that the accuracy degradation introduced by training popular CNN models on that level of precision (using 1- or 2-bit weights and feature maps) does not have a considerable effect for several real-life applications, while their reduced memory footprint offers significant space for acceleration with customised hardware. In this context, \textsc{Finn} was the first FPGA framework to undertake the challenge of optimising hardware units for BNNs. With more studies on the optimisation of FPGA-based BNNs \cite{Zhao_2017}\cite{Liang2018fpbnn} and ternary networks \cite{Boucle2017fpl}, the automated optimised mapping of binary and ternary operations can offer FPGAs a competitive advantage over competing platforms which cannot be customised to efficiently support such operations.

\textbf{\textit{Objective 4}. Integration with the existing deep learning infrastructure.}
So far, Caffe has been the best-supported framework by CNN-to-FPGA automated tools, as discussed in Section \ref{sec:inter_IN}. However, other interfaces such as TensorFlow by Google seem to gain attention by the academic and industrial communities because of the wide variety of supported machine learning models and the provided flexibility for deployment across different heterogeneous systems. While FP-DNN provides back-end support for TensorFlow, building the necessary infrastructure for newer deep learning frameworks, such as MXNet, PyTorch, Caffe2, CoreML and CNTK, and developing methodologies that can efficiently process the Intermediate Representation (IR) of each framework in order to yield optimised FPGA mappings can constitute a critical factor in exposing the deployment of CNNs on FPGAs to the wide community of deep learning researchers and practitioners.


As a recent example towards this direction, Xilinx introduced reVISION\footnote{https://www.xilinx.com/revision}, a resource suite that allows rapid development of highly responsive embedded-vision reconfigurable systems, through a software-level design flow. The reVISION stack enables the combination of machine learning and computer vision algorithms with reconfigurable devices, while allowing sensor fusion and high-level connectivity and supporting standard frameworks such as Caffe for application development.

\textbf{\textit{Objective 5}. FPGA-based CNN training.}
In both academic and industrial work, GPUs constitute the main computing platform for the acceleration of the training task. Big industrial companies such as Facebook and Baidu typically employ GPU-powered clusters, situated in data centres, to handle model training. For data centres, the power and cooling infrastructure costs constitute one of the most critical factors of the operational expenses. Since GPUs provide high performance at the expense of high power consumption, they become costly platforms to maintain. {\color{black}This fact has led Google to design and deploy the Tensor Processing Unit (TPU) ASIC \cite{Jouppi2017} in its servers for the training and inference stage of machine learning models. With next-generation FPGAs achieving promising performance and power efficiency \cite{Nurvitadhi_2017}, FPGAs can provide a high-performance, low-power alternative back end for the training task. To this end, big industrial companies such as Microsoft \cite{Caulfield2016} and Amazon\footnote{https://aws.amazon.com/ec2/instance-types/f1/} have modified their data centre facilities to host FPGAs and offer opportunities for the training of neural network models using FPGAs.} Moreover, recent advances in low-precision neural network training \cite{Zhou_16}\cite{Hubara_16}\cite{asyncsgd2017isca}\cite{Chen_17}\cite{NIPS2017terngrad} offer room for customisation and variable-precision arithmetic that suits FPGA-based computing and cannot be efficiently exploited by conventional programmable platforms. At the moment, FPGA-based CNN training has only slightly been explored \cite{Wenlai_Zhao_2016}\cite{Park2017micro} with a lot of space for further investigation.

\textbf{\textit{Objective 6}. Hardware-Network co-design.}
Ideally, a fully automated CNN framework would provide an end-to-end toolchain. Starting from a user-specified dataset and a target application, the tool would start by analysing the data and proposing an initial neural network model. By including the hardware performance and power consumption as metrics in the training phase, the hardware tunable parameters and the model weights and topology would be jointly modified during the optimisation process in order to co-optimise both the application-level accuracy and the required inference execution time and power consumption. Such a methodology would encompass the algorithmic model 
design 
together with the generation of efficient hardware under a holistic view that could potentially close the loop between CNN design and implementation. We envision frameworks that would provide this functionality as a long-term objective for the community in order to make steps towards the efficient hardware execution of high-performing neural networks.

\section{Conclusion}

This paper presents a survey of CNN-to-FPGA toolflows. A comparative analysis of their main characteristics and features indicates the strengths and weaknesses of each tooflow together with its mapping techniques. The non-uniform evaluation methodologies that have been employed so far introduce limitations in the comparison between the toolflows and fall short of taking into account both the computational and memory resources of the target FPGA platform. To this end, a comprehensive benchmark suite and thorough evaluation metrics are proposed in order to lead to further and more rapid developments in CNN-to-FPGA toolflows. Moreover, based on recent developments in deep learning, promising research directions and future objectives are identified in order to address the emerging challenges of the field, exploit FPGA-specific performance optimisations and enhance the accessibility of FPGAs to the wide community of deep learning.

\bibliographystyle{ACM-Reference-Format}
\bibliography{refs}

\end{document}